\documentclass{IEEEtran} 
\usepackage{times}
\usepackage{subcaption}
\usepackage[numbers]{natbib}
\usepackage{multicol}
\usepackage[bookmarks=true]{hyperref}
\usepackage[normalem]{ulem}

\usepackage{float}
\usepackage{placeins}

\usepackage{amsmath}
\usepackage{amssymb}
\usepackage{mathtools}
\usepackage{amsthm}

\usepackage{hyperref}
\usepackage{amssymb}  
\usepackage{algorithm} 
\usepackage{graphicx}

\graphicspath{ {figs/} }
\usepackage{float}
\usepackage{
    amsmath,
    float,
    xcolor,
    marginnote,
    subcaption,
    pgfgantt,
    times,
    outlines,
    wrapfig,
    romannum,
    csquotes,
    bm,
    algorithm,
    algpseudocode,
}

\usepackage{lipsum}
\usepackage{soul,xcolor}

\setstcolor{red}
\usepackage{tabularx}
\usepackage{adjustbox}
\usepackage{graphicx}
\usepackage{booktabs}
\usepackage{soul}

 \newcommand{\todo}[1]{{\small \color{red} \textsf{TODO: {#1}}}}

\usepackage{array}
\newcolumntype{C}[1]{>{\centering\arraybackslash}b{#1}}

\newcommand{\osnote}[1]{{}}
\newcommand{\rgnote}[1]{{}}

\newcommand{\jsnote}[1]{{}}
\newcommand{\mvnote}[1]{{}}
\newcommand{\assignTo}[1]{{}}
\newcommand{\revisit}[1]{{}}

\newcommand{\ba}{{\mathbf a}}
\newcommand{\bo}{{\mathbf o}}

\begin{document}

\pagenumbering{arabic}
\title{Multi-Stage Cable Routing through Hierarchical Imitation Learning}




\author{
Jianlan Luo$^{*,\dagger}$, \textit{Member, IEEE,}
Charles Xu$^{*}$,
Xinyang Geng,
Gilbert Feng,
Kuan Fang, \textit{Member, IEEE,}
Liam Tan,
Stefan Schaal, \textit{Fellow, IEEE}, and
Sergey Levine, \textit{Member, IEEE}
\thanks{$^{*}$equal contribution}
\thanks{$^{\dagger}$Correspondence to jianlanluo@eecs.berkeley.edu}
\thanks{Jianlan Luo, Charles Xu, Xinyang Geng, Gilbert Feng, Kuan Fang, Liam Tan, and Sergey Levine are with the Berkeley AI Research Lab (BAIR), Department of Electrical Engineering and Computer Science, University of California, Berkeley, Berkeley, CA 94720 USA.}
\thanks{Stefan Schaal is with Intrinsic Innovation LLC, Mountain View, CA 94043 USA.}%
}%



\maketitle

\begin{abstract}
We study the problem of learning to perform multi-stage robotic manipulation tasks, with applications to cable routing, where the robot must route a cable through a series of clips. This setting presents challenges representative of complex multi-stage robotic manipulation scenarios: handling deformable objects, closing the loop on visual perception, and handling extended behaviors consisting of multiple steps that must be executed successfully to complete the entire task. In such settings, learning individual primitives for each stage that succeed with a high enough rate to perform a complete temporally extended task is impractical: if each stage must be completed successfully and has a non-negligible probability of failure, the likelihood of successful completion of the entire task becomes negligible. Therefore, successful controllers for such multi-stage tasks must be able to recover from failure and compensate for imperfections in low-level controllers by smartly choosing which controllers to trigger at any given time, retrying, or taking corrective action as needed. To this end, we describe an imitation learning system that uses vision-based policies trained from demonstrations at both the lower (motor control) and the upper (sequencing) level, present a system for instantiating this method to learn the cable routing task, and perform evaluations showing great performance in generalizing to very challenging clip placement variations. Supplementary videos, datasets, and code can be found at \url{https://sites.google.com/view/cablerouting}. 
\end{abstract}

\section{Introduction}

Complex, multi-stage robotic manipulation tasks often arise in robotic manipulation applications of practical interest: from home robots that need to prepare a meal to industrial robots that need to assemble a complex device. Many of the tasks we might want to automate in these settings consist of complex low-level behaviors and also require these behaviors to be sequenced appropriately to perform the overall task. This presents a major challenge: when a robot na\"{i}vely executes a sequence of primitive behaviors to perform a complex task, the probability of failing at the task grows multiplicatively with each stage. Advances in perception, control, and robotic learning can enable each stage of a task to be more performant, but as long as sequencing stages na\"{i}vely leads to such difficulties, elaborate multi-stage tasks that consist of a sequence of individually difficult primitives will remain out of reach. In this paper, we examine how hierarchical imitation learning with levels of learned primitives and high-level sequencing can address this issue, in the context of a difficult multi-stage cable routing task. Our focus is on providing for robustness at both levels of the hierarchy, not by trying to construct perfect robotic skills that never fail, but by endowing both levels of the hierarchy with the ability to correct and recover from mistakes.
\begin{figure}[t]
    \centering
        \includegraphics[width=0.49\textwidth]{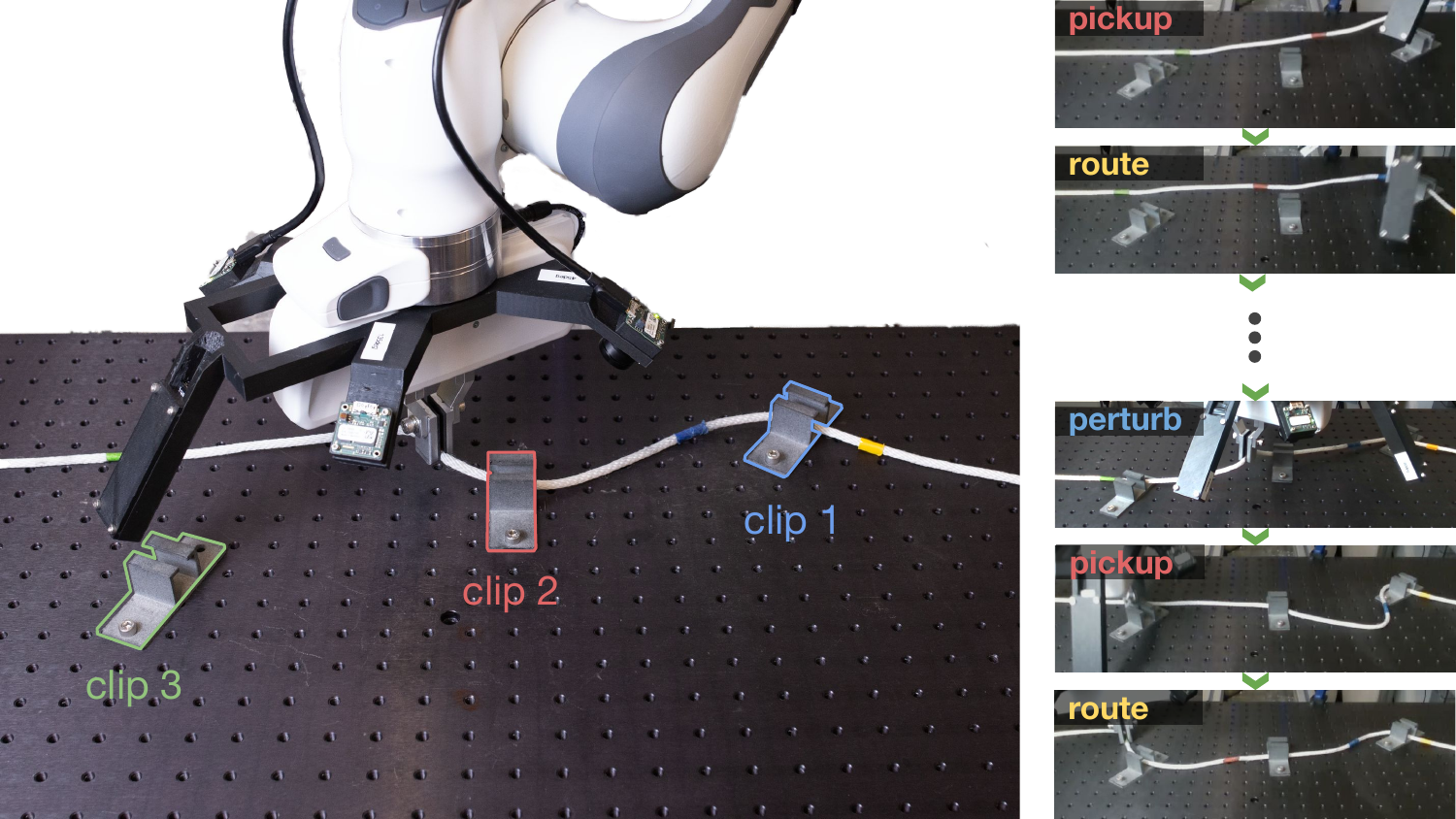}\\
    \caption{Overview of our robotic cable routing system. The robot needs to route the cable through three clips separately by executing the sequence of primitives displayed on the right.}
    \label{fig:teaser}
    \vspace{-0.5cm}
\end{figure}
We study the problem of multi-stage cable routing, where a robot routes a cable through a series of clips (see Figure~\ref{fig:teaser}). This task is representative of commonly occurring scenarios in manufacturing and maintenance, where a robot might need to route cables or hoses, and provides an interesting challenge for robotic manipulation: each stage of the cable routing task requires handling the deformable cable, reasoning about complex contact patterns between the cable and the clip, accounting for deformations, and also observing that the cable has been clipped into each clip successfully. At the same time, the higher-level sequencing of primitives might require retrying a clipping motion, securing the cable by pulling it taut, and dynamically deciding when to advance to the next clip. This task is, therefore, both practically relevant for industrial applications, and captures many of the essential characteristics of multi-stage manipulation discussed in the preceding paragraph, with challenges and ambiguity at both the lower and upper level necessitating intelligent controllers that combine perception and control, and recovery from failure.

\begin{figure*}[t]
    \centering
   \includegraphics[width=0.95\textwidth,keepaspectratio]{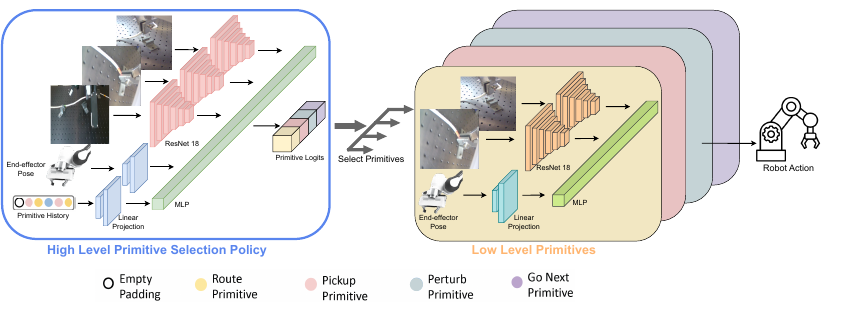}
    \caption{Overview of the hierarchical cable routing policy. The \textbf{high-level primitive selection policy} takes the robot wrist and side camera observations, as well as the history of executed primitives as input, and outputs a categorical distribution to select the next primitive. The \textbf{low-level single clip cable routing policy} only uses the wrist camera observations and the robot state to output a Gaussian distribution of robot actions. This decomposition of our system into high-level and low-level policies allows us to collect data and train policies with large flexibility, thus enabling the robot to master sophisticated cable routing tasks while reducing the overall complexity of our system.}
    \label{fig:nn_arch}
\end{figure*}
\begin{figure*}[t]
    \centering
    \includegraphics[width=0.95\textwidth,keepaspectratio]{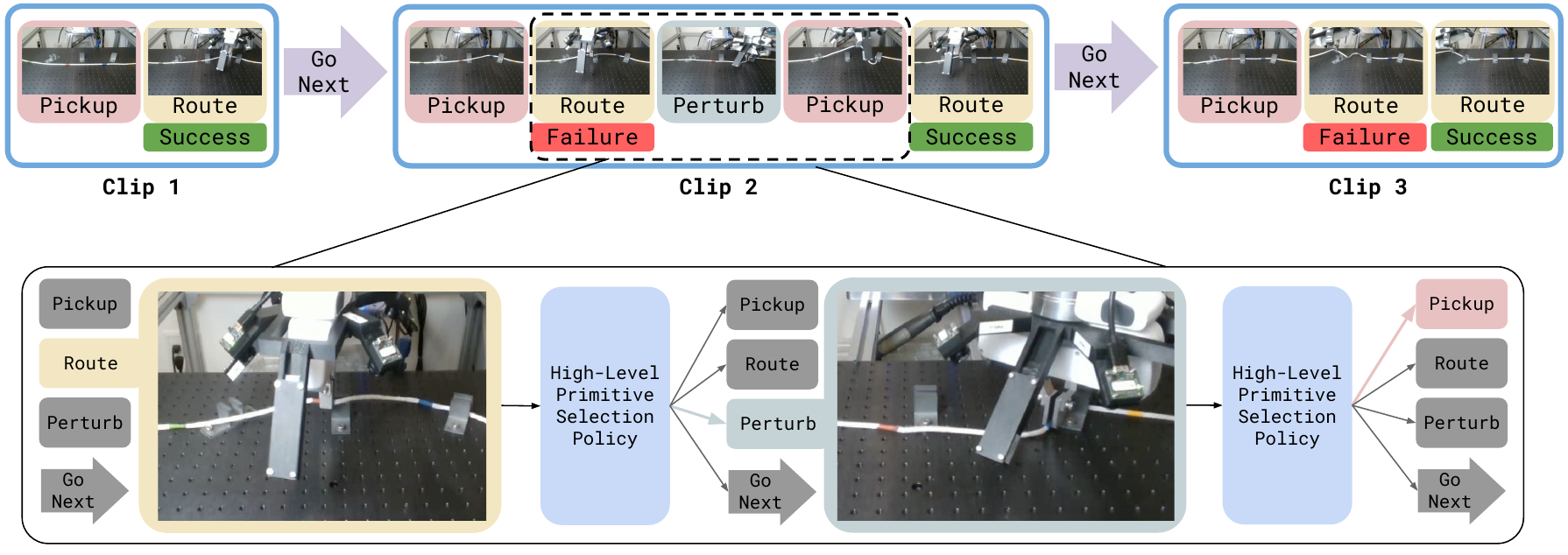}
    \caption{Sequence of primitives chained together by a successful high-level policy rollout. The primitives in the sequence are color-coded following Figure~\ref{fig:nn_arch}. After successfully routing the cable through clip 3, the policy triggers \texttt{go\_next} a third and final time, signaling the end of the trajectory.}
    \label{fig:routingseq}
    \vspace{-0.5cm}
\end{figure*}

Learning provides a powerful tool for handling complex, hard-to-model scenarios such as the manipulation of deformable cables, particularly when the robot controller needs to tightly integrate perception and control. Methods based on imitation learning (IL)~\cite{levine2018learning, brohan2022rt} and reinforcement learning (RL)~\cite{kalashnikov2018qt, kalashnikov2021mt} provide for the ability to learn vision-based controllers end-to-end from data and experience, but such methods are difficult to apply directly to complex multi-stage tasks. As we show in our experiments in Section~\ref{sec:experiments}, na\"{i}ve end-to-end learning for tasks such as the one in Figure~\ref{fig:teaser} fails even when provided with idealized demonstration data. However, learning methods can provide an excellent tool for learning basic primitives, such as inserting the cable into a single clip at a single position. Such primitives may not succeed every time, but they are relatively easy to obtain with minimal engineering effort (e.g., by using demonstration data) and can operate directly on raw image observations without any hand-designed perception system. Thus, if we can combine such primitives with an intelligent higher-level controller, we can make solving multi-stage tasks much easier. However, as mentioned previously, simply sequencing the primitives na\"{i}vely (e.g., inserting into each clip in turn) leads to exceedingly low success rates when the individual primitives are imperfect, as the probability of failing on the entire task is the product of failure probabilities at each stage. We therefore propose to employ a learned policy for higher-level primitive sequencing that can compensate for imperfections in the individual skills, resulting in a robotic learning system that is more robust than the individual parts: the higher-level can decide to automatically trigger a second attempt if the clipping failed, tighten the cable to reduce slack and select intelligently when to move on to the next clip.

Our complete system for learning-based cable routing incorporates the idea of learned controllers for recovering from failure at both levels of the hierarchy. At the lower-level, we show how we can train a clipping controller from demonstrations that can insert a cable into a clip at various orientations, respecting translational and rotational invariances, while at the same time automatically recovering from small failures through an appropriate choice of demonstration collection strategy. At the higher-level, our high-level policy selects primitives to trigger at each stage, dynamically choosing when to move on to the next clip or reduce slack on the cable. This policy is also trained with demonstrations, which are significantly easier to gather as they require the human operator to simply manually trigger one of a discrete set of primitives at each stage. We find that the process of gathering these demonstrations can be further simplified by employing an interactive imitation learning strategy, similar to DAgger~\cite{dagger}. We also explore design decisions for making this system robust and practical, including the sharing of convolutional network representations between the lower and upper layer, encoding translational and rotational invariances into the lower-level controller with an appropriate choice of view-invariant representations, and using a learned word embedding layer to encode primitive execution history.

Our primary contribution is a system for hierarchical imitation learning applied to the cable routing task. We show that learned and scripted lower-level primitives can be combined via a learned higher-level policy into a multi-stage controller that can perform temporally extended cable routing tasks, compensating for failures in the lower-level by triggering appropriate primitives at each stage. We describe the design decisions behind our system and compare our approach to methods that employ a flat end-to-end policy, as well as a na\"{i}ve scripted sequencing strategy, showing that our method not only outperforms them in terms of absolute performance but can also adapt to entirely novel situations using our finetuning mechanism with ten additional demonstrations.

\label{introduction}
\section{Related Work}

Our hierarchical imitation framework combines concepts from imitation learning and hierarchical policies to address the challenges in multi-stage cable routing. We therefore survey prior work on learning-based visuomotor control, composition of skills, and deformable object manipulation.

\noindent \textbf{Learning-based robotic control.}
Learning-based methods have been proposed for a range of robotic control problems, such as 
manipulation~\citep{pinto2016supersizing, mahler2017dex, levine2018learning, gu2017deep, levine2016end, luo2019reinforcement,zhao2022insertion}, 
navigation~\citep{shah2022gnm, zhu2017target, shah2023vint}, 
and locomotion~\citep{lee2020learning, loquercio2022learning}.
Much of the progress has been made possible by reinforcement learning (RL)~\citep{kaelbling1996reinforcement, sutton2018reinforcement, schaal2007} and imitation learning (IL)~\citep{hussein2017imitation} techniques, which enable robots to solve various tasks by learning from trial and error processes or human demonstrations. 
To solve long-horizon tasks, hierarchical RL and IL algorithms have been proposed to reuse primitive skills, reduce sample complexity, and facilitate generalization~\cite{barto2003recent, botvinick2012hierarchical, pateria2021hierarchical}.
In these approaches, the hierarchy of the task can be discovered by learning high-level and low-level policies from online interactions~\citep{Dayan1992FeudalRL, dietterich2000hierarchical, dayan1992feudal, bacon2017option, Bagaria2020OptionDU, frans2017meta, gupta2019relay} or offline dataset~\citep{le2018hierarchical, xie2020deep, krishnan2017ddco, fox2019hierarchical, kipf2019compile, shankar2020discovering, lu2021learning, wulfmeier2021data}.
Similar to several prior works on hierarchical imitation~\citep{xie2020deep, fox2019hierarchical},
our method uses a fixed set of primitives and trains a high-level model to select skill indices.  
However, we describe a set of design decisions that make it feasible to extend hierarchical imitation to a complex multi-stage cable routing task, including techniques for maximizing the generalization and invariance of a learned low-level clip insertion primitive, and strategies for data collection. We also show that our high-level policy can be fine-tuned rapidly for out-of-distribution scenarios through an interactive data collection scheme.

\noindent \textbf{Skill composition for multi-stage tasks.}
A large number of works investigate how to solve long-horizon tasks by composing primitive skills. 
With fixed control flows, behavior trees~\cite{colledanchise2018behavior, marzinotto2014towards} are widely used in robotics and control systems to switch between a finite set of tasks in a modular fashion. 
Built on pre-defined action modes and the symbolic states of the environment, Task and Motion Planning (TAMP)~\citep{garrett2021integrated, kaelbling2010hierarchical, de2013towards, garrett2020pddlstream} uses task planners to select feasible task plans and motion planners to generate trajectories for solving long-horizon tasks.
An increasing number of works use learned state estimators~\cite{migimatsu2022grounding, du2022iros, Curtis2022icra}, planners~\cite{danielczuk2019mechanical,  Wang2022GeneralizableTP, driess2021learning, Paxton2019RepresentingRT}, and skill trees~\cite{konidaris2012robot} to apply planning and sequencing to real-world environments where the ground truth states and models are not provided.
Such methods' performance usually relies on the accuracy of the state estimators and model of the environment. It's very challenging to build these components for the considered task; for instance, it’s not obvious how to build an effective state estimator for deformable cables, particularly in the presence of occlusions that tend to occur with clips and with the arm.   
In contrast to these works, our method learns both the high-level and low-level policy to solve multi-stage cable routing tasks.
Recent works also use natural language to compose hand-designed or learned skills using pre-trained large language models to leverage prior knowledge about the tasks~\citep{Ahn2022DoAI, Shridhar2022PerceiverActorAM, Jiang2022VIMAGR}. 
Our work does not utilize any additional sources of data for pre-training and performs the task only based on visual inputs.

\noindent \textbf{Deformable object manipulation.}
The manipulation of deformable materials, such as cloth, rope, and liquid, presents a major challenge for robots~\citep{hopcroft1991case, yin2021modeling}. 
Several works have designed motion planners and controllers based on models of deformable objects for specific domains~\citep{sun1996modeling, wada2001robust, hirai2001robust}.
Most of these methods rely on extensive domain expertise, and their design is typically specific to a particular type of object, material, and task.
Recently, an increasing number of data-driven methods have been proposed to train robots to manipulate deformable objects through physical interactions. 
\citep{yan2020self, luo2018deformable, huang2023act, li2020causal, grannen2020untangling} learn to estimate the configuration and model the dynamics of cloth and ropes based on visual inputs.
In \citep{li2022see, luo2018cloth, she2021cable, 9982065, 10161069}, multi-modal sensory inputs, such as haptics and audio, are utilized to enable a robot to perform challenging deformable object manipulation tasks.
These works have enabled robots to perform a wide range of tasks such as folding clothes~\citep{maitin2010cloth, lee2015learning, weng2021fabricflownet, xu2022dextairity}, rope reshaping~\citep{nair2017combining, zhu2019robotic, yan2020self, 9732654}, knot tying~\cite{wang2006knot, morita2003knot}, rope untangling~\citep{viswanath2021disentangling, viswanathautonomously}, dynamics rope manipulation~\citep{chi2022iterative, zhang2021robots}, etc.
In this paper, we are interested in the application of cable routing with potential applications in industrial tasks. 
Prior work~\citep{she2021cable, waltersson2022planning} has studied similar problems using hand-designed controllers and planners.
Many previous works on rope manipulation such as \citet{yan2020self} and \citet{Schulman2013TrackingDO} model ropes as linear objects and then estimate the best matching cable shape from perceptual inputs, which assumes full visibility of the cable in the scene and can be problematic in our setting due to occlusions introduced by clips.
As far as we know, ours is one of the first works that use learning-based methods to enable robots to route cables across multiple clips in unseen scenarios based on visual inputs.

\label{related work}
\section{System and Task Setup}\label{sec:task}

\begin{figure}[t]
    \centering
        \includegraphics[width=0.499\textwidth]{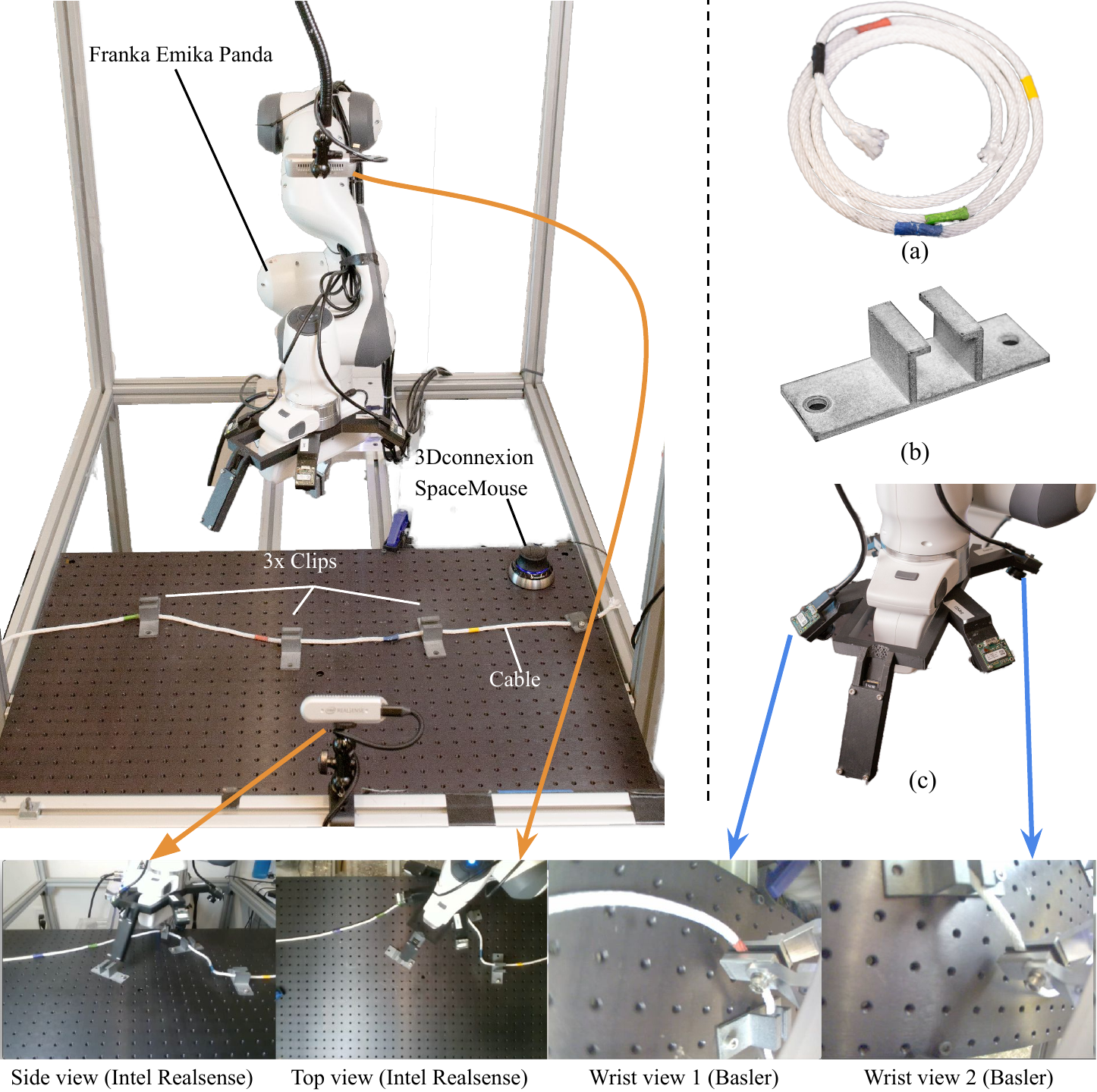}
    \caption{\textbf{Top left}: Picture of the entire system setup depicting the robot arm mounted to a frame along with a metal board where three clips are fixed. \textbf{(a)}: White 1/4$''$ thick poly cord with colored segments. \textbf{(b)}: Clip with 1cm wide opening. \textbf{(c)}: Eye-in-hand camera mounted to the wrist of the robot. \textbf{Bottom first from left}: Side view of the workspace from the RealSense camera. \textbf{Bottom second from left}: Top view of the workspace from RealSense camera. \textbf{Bottom second from right}: Wrist view 1 from Basler camera. \textbf{Bottom first from right}: Wrist view 2 from Basler camera.}
    \label{fig:system_steup}
    \vspace{-0.5cm}
\end{figure}

Figure~\ref{fig:system_steup} provides an illustration of our robotic manipulation setup for cable routing. Our system consists of one Franka Panda robot, two Basler RGB cameras mounted on the robot's end-effector, and two additional Intel RealSense cameras mounted on the work cell, one providing a top-down view and the other providing a front-facing view. The end-effector cameras provide a close-in view of the cable and facilitate robust insertion of the cable into clips, while the static cameras provide an overview of the workspace to aid in routing the cable through multiple clips while observing the resulting deformations. We only use the RGB component of the Intel RealSense output in our implementation. A space mouse is also used for collecting human demonstrations.

While one end of the cable is fixed on the table, the task is for the robot to pick up some part of the cable and then route it through all the clips present on the table sequentially, as shown in Figure~\ref{fig:teaser}. Each instance of the task has a different placement of the clips in terms of their position and orientation. 

Routing cables through clips can be difficult: since the cables are deformable, they can take an infinite variety of shapes, and their behavior is affected by the complex interaction between the gripper, the cable, and multiple other clips that the cable might already be inserted into. To correctly slot the cable into a new clip, the robot must keep track of the point on the cable that contacts the clip while grasping it at a different location, which necessarily requires closing the loop on visual perception. So it is crucial to learn a robust reactive policy that can not only perform reasonably routing a single clip but also employs recovery and retry mechanisms in terms of failure.  This makes learning-based control particularly appealing, as it can enable closing the loop on vision with minimal manually encoded domain knowledge. 



\label{problem}
\section{Cable Routing via Hierarchical Imitation Learning}\label{sec:method}

It is very difficult to learn a cable routing policy for the multi-stage task described in the previous section that never fails. Instead, we could design a hierarchical architecture where a higher-level policy compensates for the deficiency of lower level skills, sequencing appropriate primitives to retry, repositioning the cable, and dynamically deciding when to proceed to the next clip. Merely sequencing primitives na\"ively is unlikely to lead to much benefit in this case, but a strategy that selects the primitives dynamically to compensate for mistakes could lead to significantly better performance.

In this section, we present our complete system of sequential multiple-clip cable routing.
Our policy is structured in a hierarchical fashion: at the low level, we have several primitive policies that individually perform some part of the cable routing task. These primitives comprise a variety of simple scripted behaviors for picking up and moving the cable, as well as a learned clip insertion primitive that is trained to put a grasped cable into a single clip. This single clip policy, which addresses by far the most challenging of the low-level skills, is designed so as to make it maximally invariant to clip position and orientation. At the higher level, a primitive selection policy is trained to integrate these primitives together to perform the entire multi-stage routing task, using visual observations to select which primitive to trigger. This higher level policy uses a more global view of the scene to route the cable through multiple clips. Particularly unfamiliar or unusual clip configurations that differ significantly from those seen in the training data can still cause the higher level policy to fail, even when the invariant low-level skills generalize. We therefore additionally explore how the higher level policy can be finetuned efficiently via an interactive data collection procedure to adapt it to especially difficult configurations for which zero-shot generalization fails.

\subsection{Low-Level Clip Routing Policy} \label{sec:low-level}

We will first describe our low-level single clip routing policy. The higher-level policy in our system can make use of a number of primitives, the rest of which we will summarize in Section~\ref{sec:high-level}, but the most important of these is the policy that attempts to insert the cable near the current grasp point into the closest clip. This primitive performs by far the most complex low-level task, which requires carefully manipulating the cable into the clip while holding it at a different location. It must use visual feedback since proprioceptive readings provide the end-effector position but not the cable configuration. It must also generalize to a variety of clip positions and orientations, and handle a variety of cable configurations (though the higher-level policy can compensate somewhat by deciding to perturb the cable if it is in a particularly unfavorable shape). For these reasons, we use an end-to-end imitation learning approach to train this primitive. The neural network architecture for this policy is illustrated in Figure~\ref{fig:nn_arch}. In this section, we describe the training procedure and dataset for this policy, as well as techniques we employ to maximize the invariance and generalization of this policy with respect to variation in the clip position and orientation, which are based on using wrist-mounted cameras and data augmentation.

\begin{figure}[h]
    \centering
        \includegraphics[width=0.499\textwidth]{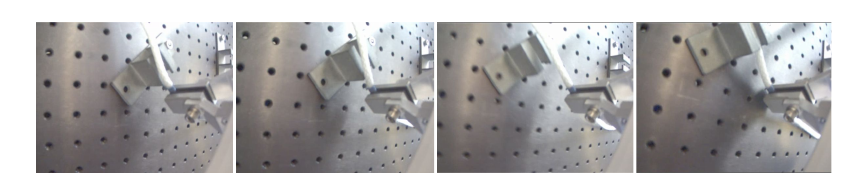}
    \caption{A recovery trial. The leftmost frame shows an initial failure state, where the rope has missed the slot and is on the side of the clip. The rest of the demo, depicted in the remaining frames, shows a recovery resulting in successful insertion into the slot.}
    \label{fig:correctiondemo}
    \vspace{-0.6cm}
\end{figure}

\noindent\textbf{End-to-end imitation learning of single clip policy.}
The single clip policy assumes that the cable has already been grasped, though it is trained to be robust to some variability in the grasp point. In the full system, the grasp will be performed by a separate primitive. The single clip policy needs to insert the cable through a single clip.
We first collect a dataset by teleoperating the robot to perform the task in various locations and for clips with various orientations. The dataset consists of 1442 trajectories, but these trajectories are relatively quick to collect, since they do not require regrasping. Each trial is less than 10 seconds in length. At the beginning of each trial, we randomize the poses of the robot's end-effector and clip.
The teleoperator records two types of demonstrations: ``successful" and ``recovery" trials. In about 800 of the trials, the demonstrator successfully inserts the cable into the clip. In about 600 of the trials, the demonstrator intentionally moves the end-effector into some failure state (e.g., a partial or failed insertion), and then demonstrates a recovery, as shown in Figure~\ref{fig:correctiondemo}. These recovery trials are intended to ensure that the learned policy is more robust to small local mistakes.
Both sets of demonstrations are combined into one dataset, which we denote $\mathcal{D} = \{(\bo, \ba)\}$, where $\bo$ is the sensor observation, which we will describe later in this section, and $\ba$ is the robot's action.
The goal of imitation learning is to find a parameterized policy $\pi_{\theta}$ that can maximize the likelihood of the current dataset $\mathcal{D}$:
\begin{align}\label{eq:bc}
\theta = \arg\max_{\theta}  \mathop{\mathbb{E}}_{\bo,\ba \sim \mathcal{D}}  \left[\log \pi_{\theta}(\ba | \bo) \right]
\end{align}
This corresponds to a standard behavioral cloning method and can be implemented with simple supervised learning methods for the architecture in Figure~\ref{fig:nn_arch}. We use Adam optimizer~\citep{adam14} with a learning rate of $3e-4$. More details can be found in Appendix~\ref{appendix3}.

\noindent \textbf{Generalization via view-invariant representations.}
The full multi-stage cable routing task requires being able to execute the clip insertion primitive (as well as other primitives) in a variety of locations depending on the placement of the clips. Therefore, it is critical to ensure that the clip insertion skill generalizes effectively over clip configurations. We found that we could significantly improve this by training the policy to act in the frame of reference of the robot's gripper with wrist cameras. This effectively puts all of the low-level policy's sensory observations and control commands into the frame of the end-effector.
We use two eye-in-hand cameras as shown in Fig.~\ref{fig:system_steup}, as they are more robust to view shifts~\cite{Luo-RSS-21, hsu2022visionbased, agarwal2022legged}. 
For the robot's proprioceptive information, such as end-effector pose,
we reference them w.r.t. the randomized reset pose of each episode.
The full sensor observation $\bo$ therefore consists of images from two wrist cameras and the TCP pose, which we find sufficient to only take the $z$ component since other spatial information can be inferred from the wrist cameras.
The control command is a 4-D Cartesian twist with full translation components
and the rotation component around the z-axis. We detail the exact procedure of calculating these quantities by doing frame transformation in the appendix~\ref{FirstAppendix}.

\noindent \textbf{Policy network architecture.}
The single clip routing policy is represented by a deep neural network, as illustrated in Figure~\ref{fig:nn_arch}, which in the end produces a Gaussian distribution over the action at each step given the current observation $\bo$. The input images from the two wrist cameras are first fed into two convolution neural networks to obtain the corresponding feature embedding vectors. We use ResNet18~\cite{he2016deep} with group normalization~\cite{wu2018group} instead of the original batch normalization~\cite{ioffe2015batch} for simplicity. 
As mentioned in the previous section, we only take the $z$ component of the end-effector pose,
which we found practically sufficient in terms of additional spatial information. We embed the $z$-coordinate
by passing it through an additional fully connected layer to get a higher-dimension embedding.
We concatenate the 2 camera embedding vectors and the $z$-coordinate embedding vector and pass the result through a 3-layer multi-layer perceptron (MLP) network to obtain the final mean and log-standard deviation of the resulting Gaussian action distribution. This model can then be trained end-to-end via maximum likelihood on the demonstration dataset, as described previously.

\begin{figure}[h]
    
    \centering
    \begin{subfigure}[b]{0.188\linewidth}
        \hspace*{-0.3cm}
        \begin{tabular}[b]{c}
        \begin{subfigure}[b]{\linewidth}
            \includegraphics[width=\linewidth]{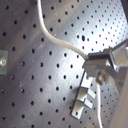}
            \label{orig45a}
        \end{subfigure}\\
        \begin{subfigure}[b]{\linewidth}
            \includegraphics[width=\linewidth]{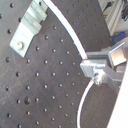}
            \label{orig225a}
        \end{subfigure}
        \end{tabular}
        \vspace*{-1cm}
        \caption*{Wrist Views}
    \end{subfigure}
    \hfill
    \begin{subfigure}[b]{0.75\linewidth}
        \hspace*{-0.3cm}
        \centering
        \begin{tabular}[b]{c}
        \begin{subfigure}[b]{\linewidth}
            \includegraphics[width=\linewidth]{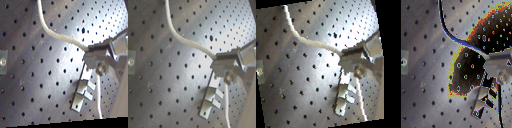}
            \label{aug45b}
        \end{subfigure}\\
        \begin{subfigure}[b]{\linewidth}
            \includegraphics[width=\linewidth]{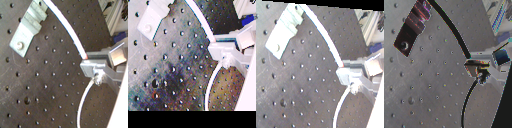}
            \label{aug225b}
        \end{subfigure}
        \end{tabular}
        \vspace*{-1cm}
        \caption*{Augmented Wrist Views after \textit{RandAugmentation}}
    \end{subfigure}
    \caption{Visualization of image augmentation. \textbf{Left}: original images from both wrist cameras when the rope is being grasped. \textbf{Right}: randomly sampled augmented images using RandAugmentation during training. Data augmentation significantly alters the images during training, enabling the policy to be more robust to distribution shift at runtime.}
    \label{fig:augmentation}
    \vspace{-0.5cm}
\end{figure}

\noindent \textbf{Data augmentation.}
We employ data augmentation techniques~\citep{ShortenK19} to facilitate the generalization capability of the learned policy to deal with image view shifts at test time such as variation in lighting conditions, small perturbations in camera pose, and variability in the objects in the scene.
In order to balance the augmentation effectiveness and the complexity of tunable hyperparameters, we employ the RandAugment~\cite{cubuk2020randaugment} technique, where a set of image transforms are chosen at random and applied sequentially each time to obtain the augmented image. Specifically, we apply two transforms sequentially and choose the strength of augmentation to be nine. We visualize the randomly augmented image in Figure~\ref{fig:augmentation}. With image augmentation, the data becomes much more diverse, allowing us to train more robust skills with smaller datasets. 

\subsection{High-Level Primitive Selection Policy}
\label{sec:high-level}

\noindent \textbf{Policy architecture overview.}
Similarly to the low-level cable routing policy, the high-level primitive selection policy is also fully parameterized by a deep neural network and trained end-to-end, as illustrated on the left side of Figure~\ref{fig:nn_arch}. However, unlike the low-level policy, which directly commands low-level actions on the robot, the high-level policy performs planning from a global perspective, and therefore we make the following design choices to suit its purpose. First, in order to choose the right clip to route, the high-level policy must be able to see the entire cable and all clips. To this end, we feed the side camera
image into an additional ResNet18 network to obtain a third embedding vector, along with the two embedding vectors from the robot wrist cameras. For sample efficiency, we reuse the pre-trained ResNet18 parameters from the low-level routing policy and freeze them for training high-level policy.
It is very important for the high-level policy to utilize a history of recent observations so that the primitive sequence for the current clip can be Markovian. For example, if a particular primitive fails, the policy might choose some corrective action, such as perturbing the cable, so that it does not fail in the exact way again. Even though the task is typically fully observed from the side camera perspective, we found that this addition improved the high-level policy's ability to correct mistakes. We, therefore, augment the input of the high-level policy with the history of at most 6 primitives chosen at prior high-level policy steps for the current clip.  We concatenate the indices of past primitives into a sequence and feed it into a learned word embedding layer~\citep{word2vec} to obtain a vector representation. We concatenate this vector with the embedding vectors of the three cameras and feed the result into a three-layer MLP to obtain the final logits for choosing the primitive.

\noindent \textbf{Primitives.} The full set of primitives available to the high-level policy consists of the learned clip insertion primitive described previously, as well as three scripted primitives: \texttt{Pickup}, \texttt{Perturb\_cable}, and \texttt{Go\_next}. We found that these three additional primitives could be solved via existing robotic solutions and did not require learning, as they perform relatively simple tasks, and the high-level policy could reliably compensate for their imperfections with appropriate choices at the upper layer of the hierarchy. Note that our hierarchical system is designed to be modular, so any of these primitives can be replaced with other solutions as well.

The clip configuration can be fully represented by a list containing the estimated
position of each clip in the desired order, which can be estimated with a standard computer vision detector. We use these estimated positions to parameterize primitives, though they are not visible to the routing policy itself.
The high-level policy proceeds through each clip in the list in order. At each clip, it can trigger a variety of primitives, and select when to advance to the next clip (incrementing the current clip index).
All of the primitives operate on the current clip that the high-level policy is handling, and the history of previously selected primitives is reset each time the high-level policy advances to the next clip.


\noindent \texttt{Pickup:} The \texttt{Pickup} primitive is used for picking up the cable at a particular position
and holding it within the gripper. We designed this separate \texttt{Pickup} primitive instead of hard-coding it as a precursor to \texttt{Route} to allow re-trying the \texttt{Route} without having to drop and re-grasp the cable.
As shown in Fig.~\ref{fig:system_steup}, different segments of the cable are marked with different colors. We move the arm out of the way to avoid occlusion before using the top camera as shown in Fig.~\ref{fig:system_steup} to detect the colored marker corresponding to the current clip to route, and then choose the center point of the detected region to construct a pick-up pose in the camera frame.
This pose is then converted back to the robot's base frame by using the top camera's extrinsic calibration information and a fixed z-position based on table height rather than the depth map from the RGBD camera because we find it to be inaccurate in practice for our thin cable. Note that this is the only camera that needs to be calibrated.



\begin{figure}[h]
    \centering
    \begin{tabular}[t]{cc}
    \begin{subfigure}[b]{0.45\linewidth}
        \includegraphics[width=\linewidth]{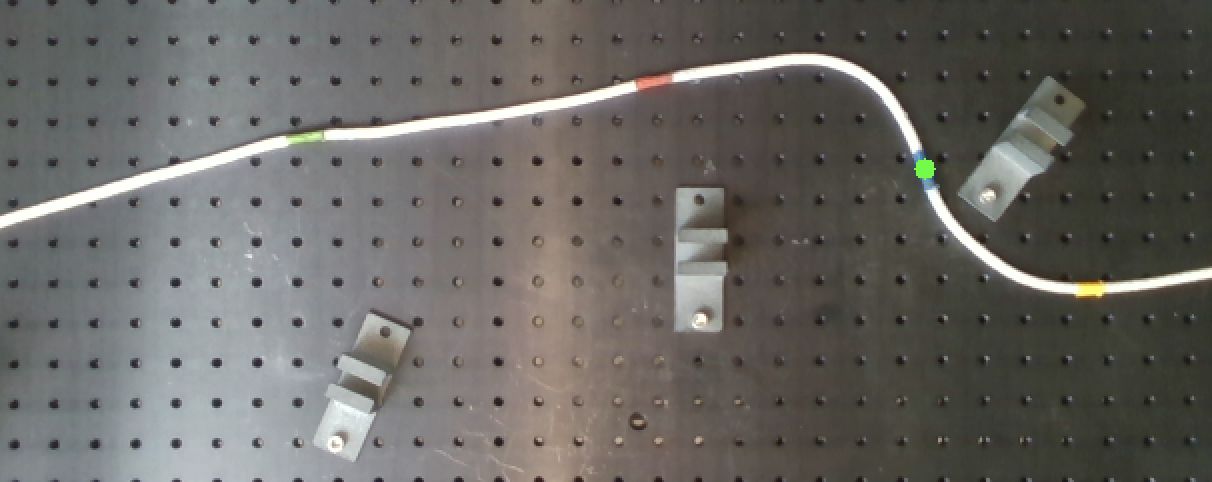}
        \caption{Pickup detection}\label{pickupdetect}
    \end{subfigure} &   
    \begin{subfigure}[b]{0.4\linewidth}
        \includegraphics[width=\linewidth]{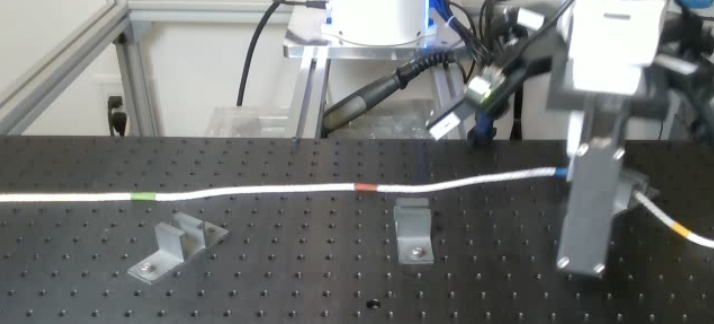}
        \caption{\texttt{Pickup}}\label{pickupimg}
    \end{subfigure} \rule[-1.2ex]{0pt}{0pt}\\
    \end{tabular}
    \caption{The \texttt{pickup} primitive uses image segmentation to detect the grasping point illustrated by the green dot in \ref{pickupdetect} and picks up the rope.}
    \label{fig:pickupcable}
\end{figure}

\noindent \texttt{Perturb\_cable:} Repeated application of various primitives can cause the cable to end up in a shape from which other primitives will consistently fail. One of the benefits of our hierarchical design is that the high-level policy can detect this by leveraging the history input and camera observations. In this case, the \texttt{Perturb\_cable} primitive can be used to rearrange the cable into a new shape, which can get it out of a pathologically difficult configuration.
We use the same detection method as described for the \texttt{Pickup} primitive to grasp the cable at specific locations, and then apply pre-defined motions to change the shape of the cable before releasing it. Although picking the right perturbation motion w.r.t. any given cable shape is generally difficult, we found in practice it's sufficient to just stretch the cable along one direction so the amount of slackness in the cable can be reduced.

\begin{figure}[h]
    \centering
    \begin{tabular}[t]{cc}
    \begin{subfigure}[b]{0.45\linewidth}
        \includegraphics[width=\linewidth]{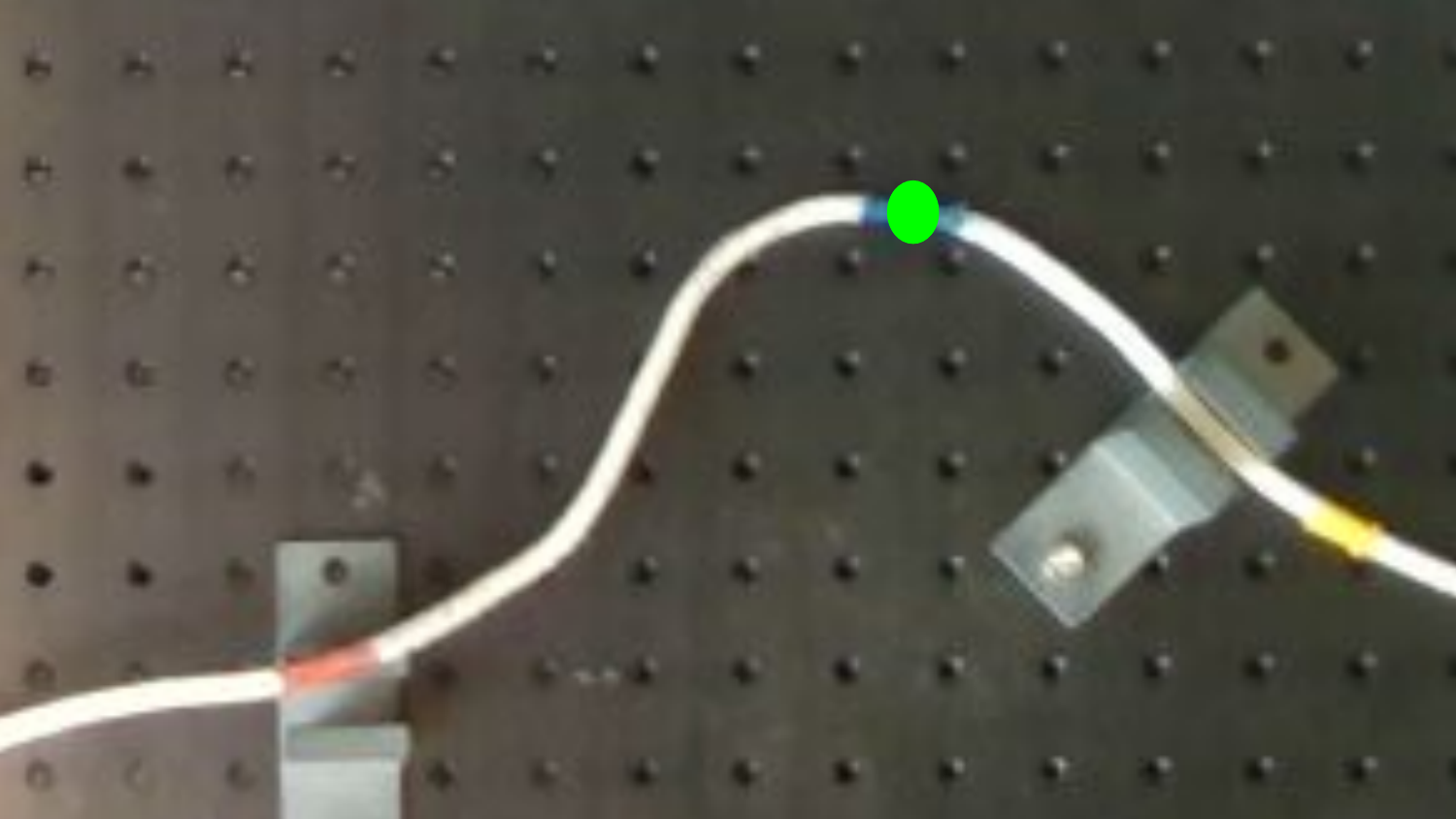}
        \caption{Before \texttt{Perturb\_cable}}\label{perturbbefore}
    \end{subfigure} &   
    \begin{subfigure}[b]{0.45\linewidth}
        \includegraphics[width=\linewidth]{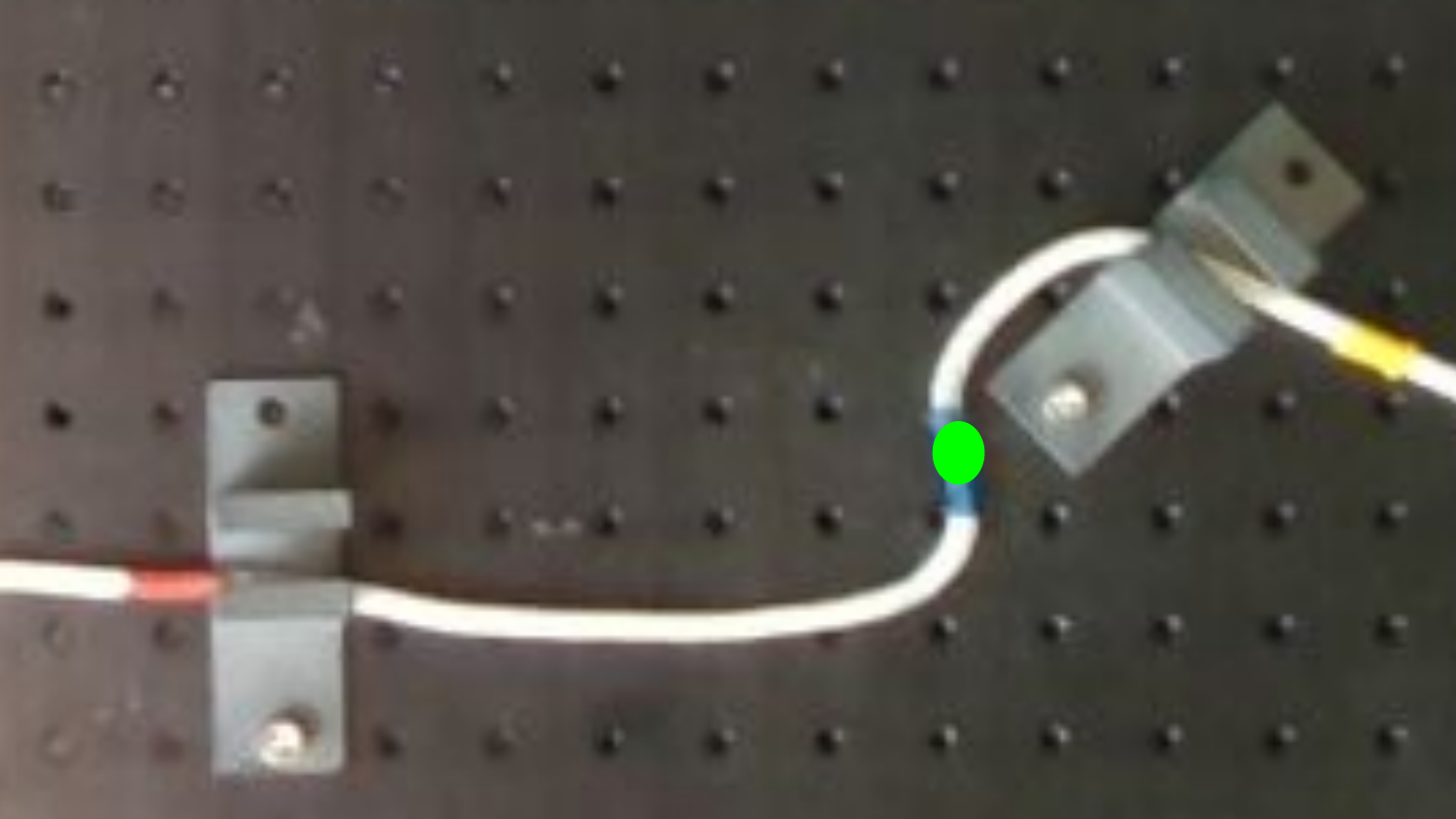}
        \caption{After \texttt{Perturb\_cable}}\label{perturbafter}
    \end{subfigure} \rule[-1.2ex]{0pt}{0pt}\\
    \end{tabular}
    \caption{\texttt{Perturb\_cable} applies a pre-defined motion to a specified point on the cable, changing the cable shape from \ref{perturbbefore} to \ref{perturbafter}. }
    \label{fig:perturbcable}
    \vspace{-0.3cm}
\end{figure}

\noindent \texttt{Go\_next:} The \texttt{Go\_next} primitive advances the task to route the cable into the next clip. 
This primitive releases the cable and moves the robot's end-effector to a position that is close to that of the next clip in the sequence, applying a small random perturbation to offset the end-effector from the clip. The random offset is sampled from the same distribution that we use to determine initial states for training the single-clip policy, as described in Sec.~\ref{sec:low-level}. That way the end-effector pose ends up within the distribution of relative poses seen during training of the low-level single-clip skill. Note that this primitive moves the end-effector while the cable is grasped in hand.

\noindent \textbf{Training the high-level policy.} We now discuss how we train the high-level primitive selection policy.
Denote images from three cameras (the two wrist cameras and the side camera) as $I_1, I_2, I_3$. We pass these images through the ResNet18 encoders and get three embeddings $e_1, e_2, e_3$. Let $h$ be the list containing the indices of up to six primitives that have been executed so far for the current clip, with the number 0 as paddings to fill the empty slots if the length of the history is less than six. Each element is a categorical variant that takes on one of four values (the learned single-clip skill and the three scripted primitives). The high-level policy $\pi_{\phi}$ takes these image embeddings and primitive history as inputs, together with the $z$ component of the end-effector pose. It outputs a categorical distribution over primitives, represented as a four-way softmax. That is,
\begin{align}
\pi_{\phi}(\cdot |e_1,e_2,e_3,h,z) \sim  \texttt{Categorical}_{4}(\cdot)
\end{align}
To train this policy, we also collect a dataset of about 250 trajectories from a human demonstrator selecting the primitives to route a cable through the clips, using a keyboard. To make the learned high-level policy more robust to local variations when making a decision, we further augment the dataset by relabelling the nearby frames where the policy needs the select primitives. Specifically, we label the adjacent three to five frames with the same primitive selection choice from the human. 
This dataset can be significantly smaller than the one we use for the low-level policy, since the image encoders are reused, and the action space of the high-level policy is significantly simpler. 
This dataset is used to train the high-level policy via a standard maximum likelihood behavioral cloning loss (i.e., a cross-entropy loss, since the output is categorical). A sample rollout of the policy is visualized in Figure~\ref{fig:routingseq}.


\subsection{Interactive Finetuning}  \label{sec:finetuning}
\begin{figure}[t]
\centering\includegraphics[width=0.95\linewidth]{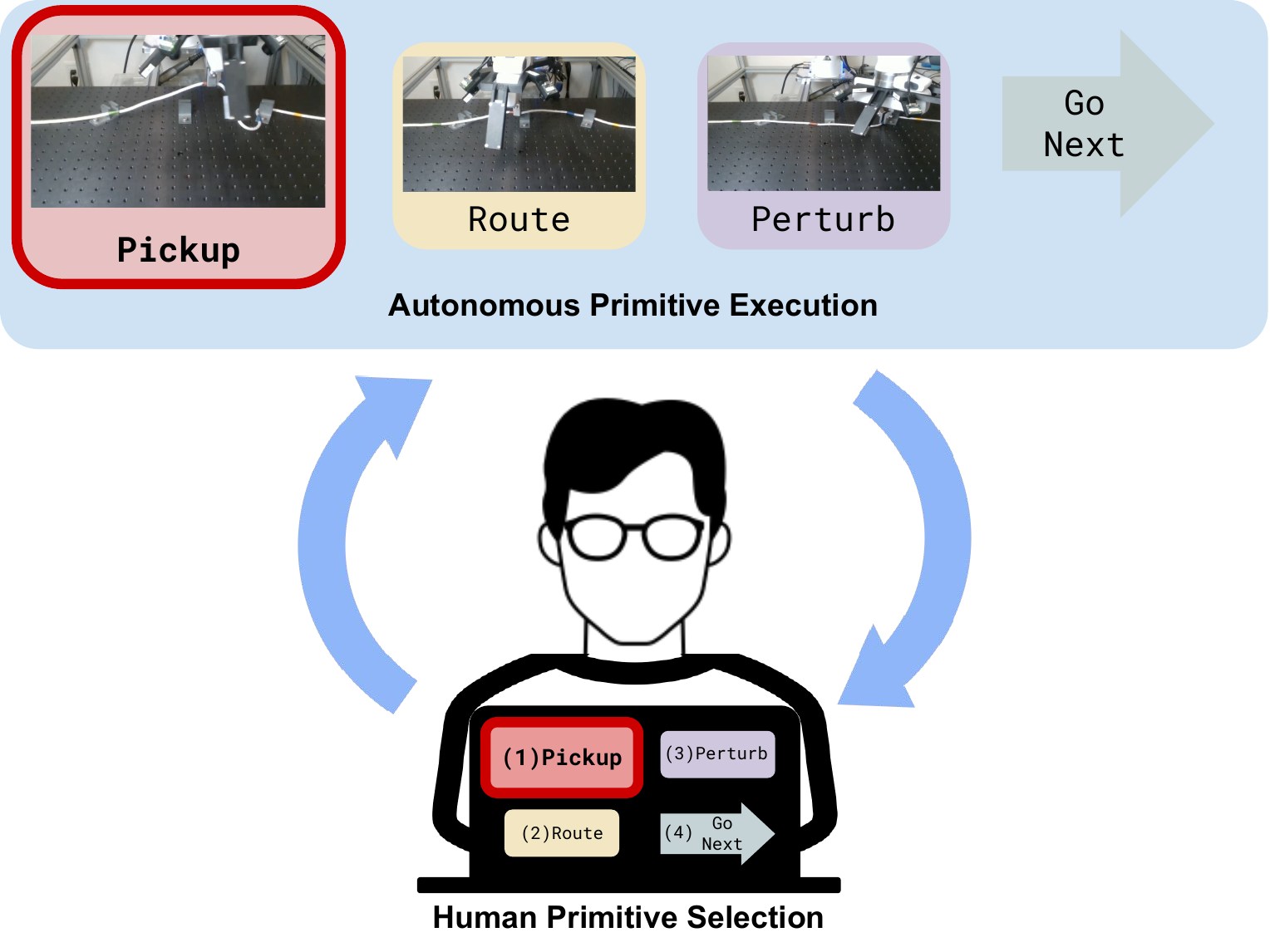}
    \caption{After the low-level primitives are acquired, data for the high-level policy can be collected by a human selecting the appropriate primitive to execute after the previous has autonomously finished until the cable is successfully routed through all the clips.}
    \label{fig:highlevel data}
    \vspace{-0.5cm}
\end{figure}

While our system provides a robust framework for tackling the cable routing problem with a variable number of clips in different locations, it is still possible for these pre-trained policies to fail to generalize to a novel condition, such as an arrangement of clips that is outside the range of clip placements seen in the training data, or a new number of clips. Additionally, it's often very desirable to have a high success rate for such systems to be deployed in real industrial settings, and it could be entirely possible that our system can't meet these stringent requirements of reliability initially. 
In these scenarios, we can further finetune our policy for better performance with a small amount of interactive training for the high-level policy. Specifically, we adopt the HG-Dagger~\cite{kelly2019hg} method, where the high-level policy attempts to complete the task under human supervision. When the policy makes a mistake and selects the wrong primitive, the human operator will intervene by overriding the high-level policy with the correct primitive, repeating this process until the episode is completed successfully. 
This interactive training procedure is easy and less disruptive since a human can immediately tell if a policy's selection will make sense without having the robot actually execute it.
After the policy outputs a primitive index, this integer number will be displayed on the computer screen waiting for human input; if the current selection will result in a non-recoverable state, the human will override the policy's selection; otherwise, we'll advance the policy's selection.
For instance, if the policy decides to skip routing the current cable and go to the next one, the human operator can override this command right after seeing it from the computer screen. We illustrate such a procedure in Fig.~\ref{fig:highlevel data}.

In our interactive finetuning experiments, we use this procedure to collect up to ten additional trials with human corrections for each new scenario. Note that we don't use the previous dataset collected for the high-level policy when finetuning to a new configuration. Rather, we finetune the weights of the MLP solely on this newly collected dataset with a learning rate warm-up. Further details can be found in Appendix~\ref{appendix3}.

\label{methods}
\section{Data Collection and Open-Source Release}
In this section, we detail our data collection procedure for both the low-level policy and high-level policy as well as the now publicly-available dataset resulting from this research.

\subsection{Data Collection}
\begin{figure}[h]
\centering\includegraphics[width=\linewidth]{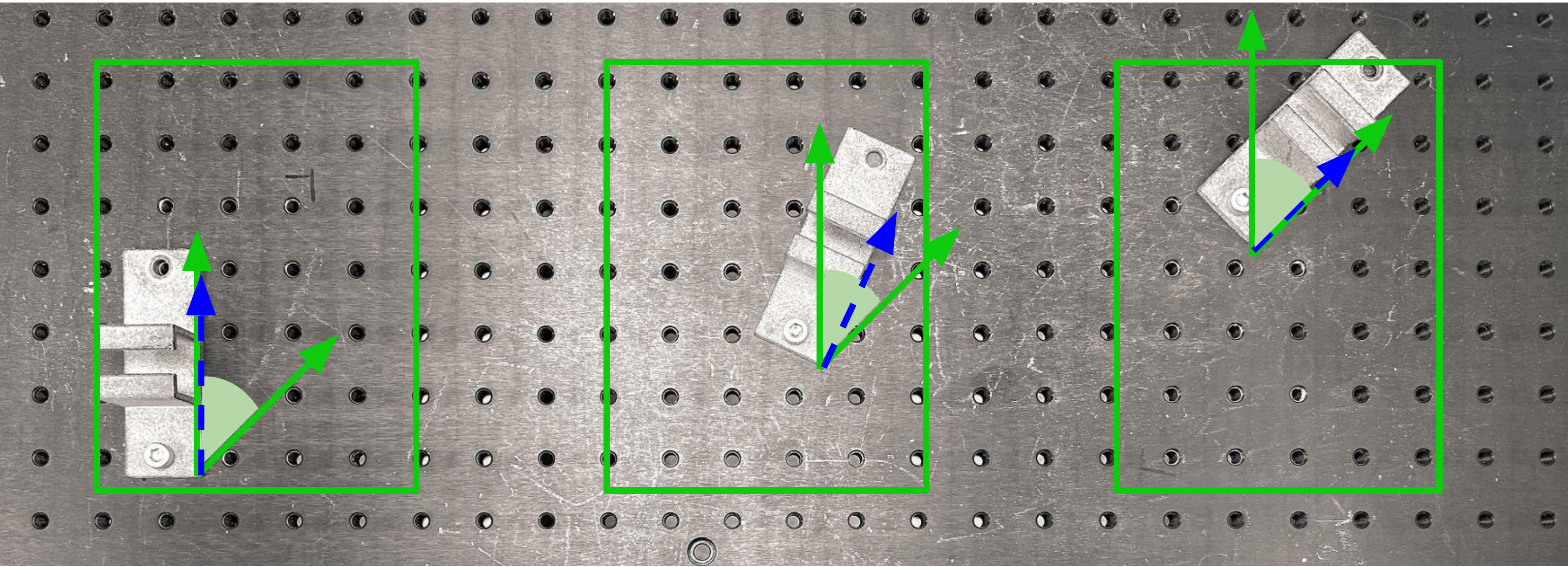}
    \caption{Each of the three clips is placed randomly within a 12.5 cm by 15 cm rectangular region, with the regions spaced 7.5 cm apart. Additionally, each clip can be rotated between 0 and 45 degrees.}
    \label{fig:clip area}
    \vspace{-0.3cm}
\end{figure}

For all of the experiments, we fixed one end of a cable to the table and fix the clips in the randomization areas defined in Fig.~\ref{fig:clip area}.

\noindent \textbf{Single clip cable routing data collection.}
For each of the three clips, we vary its position within the constraints described in Fig. \ref{fig:clip area} and collect a total of 1442 demonstration trajectories via a human expert teleoperating the robot at 5Hz. As mentioned in Sec.~\ref{sec:low-level}, of the 1442 demonstrations, about 800 start within a region in space, measuring 10cm by 10cm by 2cm in the x, y, and z direction, centered 10cm above and 5cm in front of the clip. The other 600 or so demonstrations start on the lower end of the table, where the starting position was not precisely controlled to demonstrate recovery.

\noindent \textbf{High-level data collection.}
After training the single clip routing policy and the other primitives, we collect demonstrations for the high-level policy by having a human expert trigger primitives in sequence to perform the complete multi-stage cable routing task.
We ask humans to use combinations of primitives to route cables when there are one clip, two clips, or three clips on the table. 
In each demonstration, the cable starts lying flat on the table in an arbitrary shape, and the expert inputs the next primitive for the robot to execute until the task is complete.
We record the sensory information of entire trajectories when executing low-level primitives. We also augment the dataset by labeling the adjacent states of the actual state where the human makes a selection with the same executed primitive index. 

\subsection{Open-Source Dataset}
To facilitate the reproducibility of our work, we release
the datasets used to train the low-level and high-level policies hosted on our website: \url{https://sites.google.com/view/cablerouting}. The dataset used to train the low-level policy consists of human teleoperated robot cable routing trajectories. Each trajectory contains around 20 time steps, and each time step contains a tuple of four robot camera images, robot configuration state vector, and the human teleoperator's commanded action. The entire low-level policy dataset contains 1442 such trajectories, each trajectory is around 3-5 seconds long. 
The high-level policy data consists of high-level trajectories of robot observations between primitive executions, where one timestep in a trajectory corresponds to one observation and the index human selected primitive. This dataset contains 257 such trajectories, where one full such trajectory is roughly 1 minute.
Furthermore, to ensure that our robot setup can be reproduced, we also release the CAD file of the plastic clips we used, which can be easily produced by common 3D printers.

\label{data}
\section{Experimental Evaluation} \label{sec:experiments}





In this section, we describe the experiments we conducted to evaluate our hierarchical imitation learning system for cable routing. 
One central premise of our method is the necessity of adopting hierarchical structures. To examine this claim, one natural question to ask is: can such long-horizon behavior be acquired via ``flat" imitation learning approaches? 
Hierarchical methods will be much more convincing choices if they can outperform recent state-of-the-art ``flat" methods~\citep{shafiullah2022behavior, zhao2023learning}. 
Another goal of the experiment is to validate the effectiveness of the main design choices in our system. Dissecting these design choices will facilitate the improvement and adoption of such a learning-based system in dealing with deformable objects in a much more general sense.
Finally, to validate the capability of our system, it's crucial to examine its performance in terms of generalization, both in terms of zero-shot performance and in terms of fine-tuning with a small amount of human guidance.
With these considerations in mind, our experiments study the following questions:
\begin{itemize}
    \item How effective is our low-level clip insertion policy compared to baseline approaches, and how important are the specific design choices we made in the training process for this policy?
    \item How effective is our method compared to other imitation learning methods that compose low-level skills in different ways?
    \item How well does our high-level policy generalize to novel clip arrangements?
    \item How efficient is our finetuning scheme in quickly handling out-of-distribution clip arrangements as well as improving the system to achieve desirable performance?
\end{itemize}


\subsection{Experiment Setup}
We evaluate the low-level and high-level policies under several different scenarios. 
For the trained low-level policy, we test its performance in the case of single-clip routing with different clip placement variations. 
For that, we ask the robot to repeatedly execute the trained neural network control policy autonomously, starting from various initial arm configurations. 
The high-level policy is evaluated on multiple runs of full cable routing tasks, and each such trial is only marked as successful if the cable is routed through all of the clips correctly.



\subsection{Evaluating the Low-Level Single Clip Routing Policy}

\begin{figure}[h]
    \centering
    \begin{tabular}[t]{cc}
    \begin{subfigure}[b]{0.4\linewidth}
        \centering
        \includegraphics[width=\linewidth]{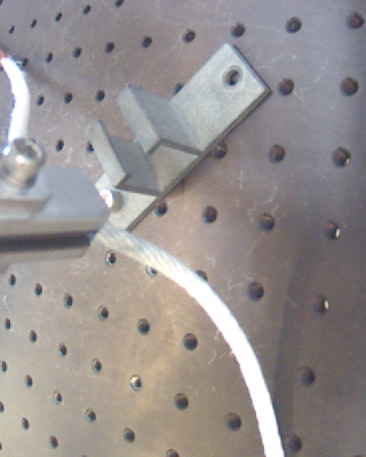}
        \caption{Curving toward clip}\label{cable_easy}
    \end{subfigure} &   
    \begin{subfigure}[b]{0.4\linewidth}
        \centering
        \includegraphics[width=\linewidth]{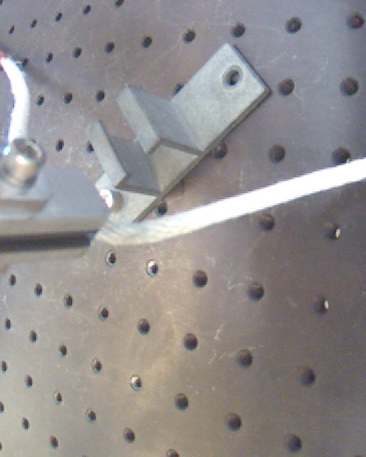}
        \caption{Curving away from clip}\label{cable_hard}
    \end{subfigure} \rule[-1.2ex]{0pt}{0pt}\\
    \end{tabular}
    \caption{Wrist view of the two different cable shapes for the same clip. \ref{cable_easy} is an easier configuration to route than \ref{cable_hard} because the cable makes a smaller angle with the opening of the clip and is easier to align.}
    \label{fig:cable_shape}
\end{figure}

We evaluate the low-level routing policies by first sampling five clip positions from the same distribution as used for training. For each of the clip positions, we identify two different cable shapes: one that curves toward the clip opening and one that curves away from it. From experience, the former is much easier than the latter for both learned policies and human experts. In the hard configurations, the cable tends to make a small radius curve of close to 90{\textdegree} near the grasping point perpendicular to the direction of the narrow straight opening on the clips. Without another arm to manipulate the curvature of the cable, this configuration makes the cable much harder to be routed through for both learned policies and human experts. An example of the different shapes is shown in \ref{fig:cable_shape}. For each combination of clip position and cable shape, we roll out each policy five times, for a total of 50 attempts per policy. Before the start of each rollout, we position the end-effector approximately 5 centimeters in front of the clip with the cable grasped in the fingers. We roll out the policy for a total of 50 timesteps and consider the trials successful if a section of the cable is completely in the clip. 

Our learned low-level policy successfully inserts the cable into the clip on 18 out of 25 trials with the easier cable shape, and 5 out of 25 trials with the harder cable shape across different clip positions, for an overall success rate of 46\%. This is a reasonable number as compared to that of a human operator: during data collection, we filter out failed episodes, but roughly a human can achieve a 60\% success rate by teleoperating the robot. While the learned policy is far from perfect, we will see below that it drastically outperforms a scripted alternative and various ablations, though it still requires an intelligent higher-level policy to compensate for its failures and mistakes.

\begin{table}[h]
    \centering
    \begin{tabular}{l|cc|c}\toprule
         Method & Easy & Hard & \textit{Overall} \\\midrule
         Scripted & 7 / 25 & 3 / 25 & \textit{10 / 50}\\
         Ours & \textbf{18 / 25} & \textbf{5 / 25} & \textbf{\textit{23 / 50}} \\\bottomrule
    \end{tabular}
    \caption{\footnotesize{\textbf{Low-level single-clip policy} comparison against a scripted baseline. Our learned policy outperforms the hand-designed alternative.}}
    \label{LowLevelScripted}
    \vspace{-0.3cm}
\end{table}

\noindent \textbf{Comparison to a scripted policy.}
In principle, a central benefit of training the low-level clip insertion policy end to end is that it can close the loop on visual perception and observe the cable as it is being inserted into the clip, in contrast to simple scripted strategies that blindly match end-effector offsets from the clip. To evaluate whether we improve on such simple baselines, we compare our low-level policy to a scripted strategy that uses the ground-truth clip position and orientation, which is not available in our vision system but does not attempt to perceive the cable itself. This scripted strategy follows a series of predefined waypoints relative to the clip to attempt to insert the cable and then wiggles the cable (using normally distributed noise added to target positions) to attempt to insert it.
The results of this comparison are shown in Table~\ref{LowLevelScripted}. Although the scripted policy even uses privileged information that is otherwise not available to our policy, we can see our policy has twice the success rate. This confirms that the single clip insertion task is challenging, and also suggests that the additional visual perception enabled by our end-to-end policy is quite helpful.

\noindent \textbf{Ablation experiments.}
To examine the effectiveness of the design choices we made in Sec.~\ref{sec:method}, we conduct extensive ablation experiments on the low-level policy. The results are shown in Table~\ref{LowLevelAblation}. We first note the drastic drop in performance when we omit image augmentation, highlighting its utility in facilitating robustness. 
Additionally, using separate ResNet18 encoders for the two eye-in-hand cameras in the low-level policy provides marginal benefit. This is reasonable and suggests that the network might have enough capacity to digest current inputs.
Replacing the eye-in-hand cameras with the external side camera view yielded a drastic drop in performance, indicating the importance of view-invariant representations. Finally, excluding the correction demonstrations from the offline dataset caused a noticeable drop in performance, with many of the failures involving the cable missing the clip and the policy failing to lift up the cable to recover and retry routing within the 50 rollout timesteps.


\begin{table}[h]
    \centering
    \begin{tabular}{l|cc|c}\toprule
         Design Choice & Easy & Hard & \textit{Overall} \\\midrule
         No image augmentation & 9 / 25 & 2 / 25 & \textit{11 / 50}\\
         No shared ResNet & \textbf{18 / 25} & 3 / 25 & \textit{21 / 50}\\
         Side camera view & 5 / 25 & 1 / 25 & \textit{6 / 50}\\
         No correction data & 15 / 25 & 3 / 25 & \textit{18 / 50}\\
         \hline
         Our full method & \textbf{18 / 25} & \textbf{5 / 25} & \textbf{\textit{23 / 50}} \\\bottomrule
    \end{tabular}
    \caption{\footnotesize{\textbf{Ablation experiments for the low-level policy}, showing performance on single clip insertion after ablating each design choice. The variant with decoupled ResNet encoders performs somewhat similarly, but the other variants are significantly worse than our full design.}}
    \label{LowLevelAblation}
    \vspace{-0.5cm}
\end{table}
\subsection{Evaluating the High-Level Policy for Cable Routing}

We first evaluate our full hierarchical system on one-, two-, and three-clip routing tasks. For a given number of clips, we sample each clip position from the same distribution as used for our training data. For each number of clips, we evaluate four randomly sampled configurations and six trials each, for a total of 24 trials. At the start of each trial, we place the cable flat on the table. Then we roll out the high-level policy by executing each primitive that it outputs until the policy outputs \texttt{go\_next} at the last clip. We consider the trial successful only if the cable has been routed through all of the clips that are currently on the table.

\begin{table}[h]
    \centering
    \begin{tabular}{l|cccc}\toprule
         & One Clip & Two Clips & Three Clips & \textbf{Total} \\\midrule
         Success Rate & 19 / 24 & 14 / 24 & 12 / 24 & \textbf{45/72}\\\bottomrule
    \end{tabular}
    
    \caption{\footnotesize{\textbf{Hierarchical policy evaluation} for different numbers of clips, evaluating in-distribution test scenarios.}}
    \label{GeneralEval}
\end{table}

\begin{figure}[h]
    \centering 
    \includegraphics[width=0.95\linewidth]{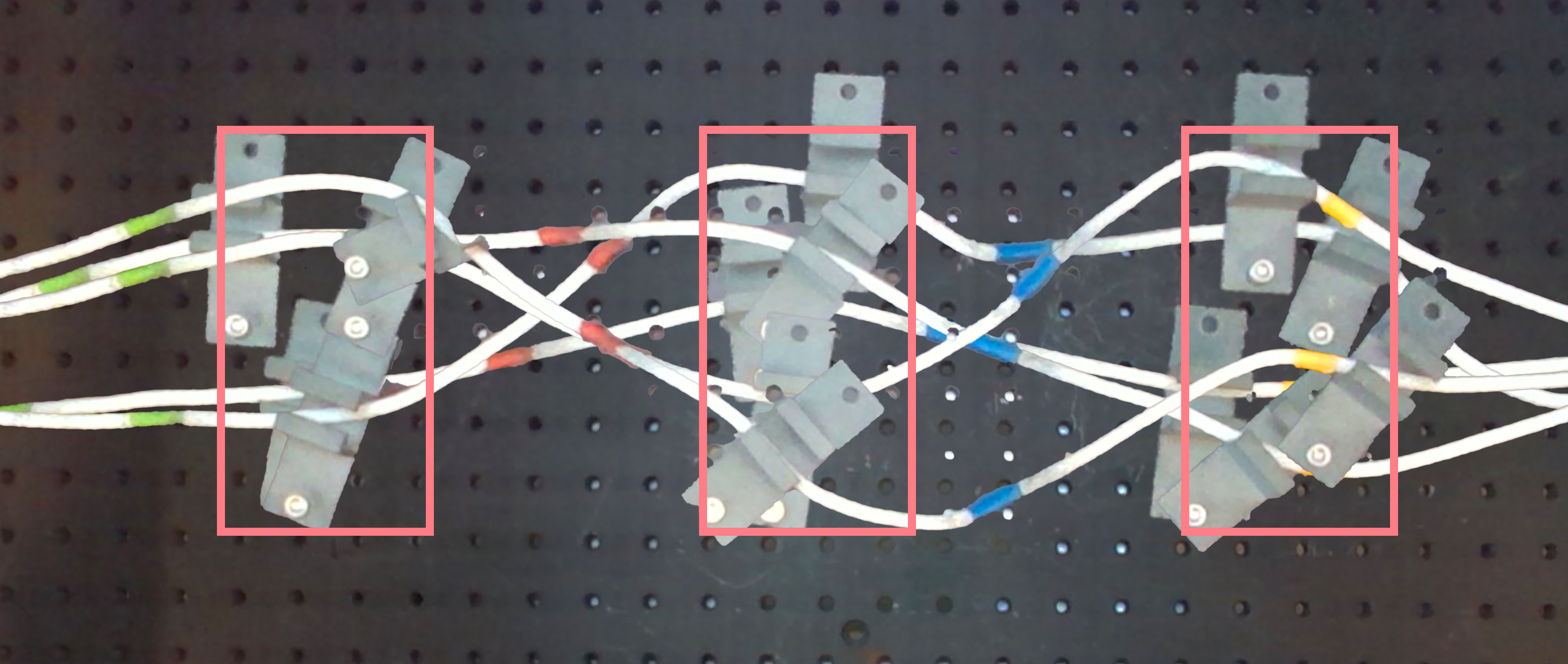}
    \caption{The red boxes depict the randomization areas where each of the three clips could be placed independently from one another, creating large variations in the subsequent cable shape.}
    \label{fig:distribution}
\end{figure}

The results are presented in Table~\ref{GeneralEval}. Fig.~\ref{fig:distribution} gives a visualization of how we conduct this randomization procedure. It's important to note that the performance of our system doesn't drop exponentially as the number of clips increases from one to three. Rather, it maintains a reasonable performance in the most challenging three-clip scenario compared to the one-clip case; though the task there is much harder. This resonates with our key motivation that the smart combinatorial use of primitives can compensate for the deficiency of each individual one; so that the overall performance of the resulting system will largely not be subject to compounding errors in long-horizon tasks in a mechanical way. 



\noindent \textbf{Baseline comparison.} Table~\ref{HighLevelBaselines} compares the experimental results of our model with the following baselines:
\begin{enumerate}
    \item State machine: The primitives are composed together using a state machine that sequentially performs \texttt{pickup}, attempts to \texttt{route} through the current clip twice before prompting \texttt{go\_next}, and then executes \texttt{perturb\_cable} before the \texttt{pickup} for the next clip. This sequential composition was inspired by strategies that the authors found to work reasonably efficiently at successfully completing the routing task during demo collection.
    Note that this strategy is not as na\"{i}ve as simply executing the low-level clip insertion policy repeatedly, and actually includes a scripted strategy for correcting mistakes.
    While it is indeed possible to always develop a better state machine, we found it increasingly complex in dealing with combinatorial variations; eventually became infeasible under practical constraints. 
    \item Flat BC policy: To verify the necessity of hierarchical structures, we train a flat end-to-end BC policy on the long-horizon trajectories during the high-level policy data collection phase. We adopt the same policy neural network architecture as in Fig.~\ref{fig:nn_arch}, with an additional action dimension to control the gripper closure. We found it achieved 0\% success rate out of 24 trials; and it never succeeded even in routing the first clip, which corresponded to failure modes such as not picking up the cable, directly moving the arm out of the board, missing the clip, prematurely dropping the cable before it's routed, and combinatorial of such. This indeed suggests the proposed task is a challenging one that poses difficulties in a compounded way that necessitates reasonable hierarchical approaches dealing with those difficulties modularly.
    \item Behavior Transformer(BeT): One possible reason that flat BC policy performs poorly is the multimodal nature of human demonstrations. To address this point so that we can make concrete a conclusion in terms of hierarchical structure. We compare to a BeT policy~\citep{shafiullah2022behavior}, a promising approach specifically geared towards handling multi-modal inputs, trained on the same dataset we train the flat BC policy as mentioned in the last paragraph. We refer to Appx.~\ref{appendix3} for implementation details of BeT. It is observed that BeT policy also ended up with 0\% success rate as well as exhibiting similar failure modes as the flat BC policy, including not grasping the cable and missing the clip. This suggested although human demonstrations may be naturally multi-modal, the bottleneck of the proposed problem is largely orthogonal to that. 
    \item Action Chunking(ACT): Results from the aforementioned flat BC policies motivate the use of hierarchical approaches to address error-compounding issues in this long-horizon task. One natural intermediate step between a fully flat method to a fully hierarchical method such as ours with complex machinery is the semi-hierarchical approach. One instance of such an approach is Action Chunking(ACT)~\citep{zhao2023learning}, which employs multiple-step action prediction and ensembles to alleviate issues of error accumulation. We detail the algorithm implementation in Appx.~\ref{appendix3}. While we did find the executed action smoothness improved, the trained ACT policy was not able to succeed at the task at all. The failure modes are missing the clip and dropping the cable to pick up the next section when the current clip is not routed.
\end{enumerate}

\begin{table}[t]
    \centering
    \begin{tabular}{l|c}\toprule
         Model & Success Rate \\\midrule
         State Machine & 5 / 24\\
         End-to-End BC & 0 / 24\\
         End-to-End BeT & 0 / 24\\
         End-to-End ACT & 0 / 24\\
         Hierarchical Imitation (ours) & \textbf{12 / 24} \\\bottomrule
    \end{tabular}
    \caption{\footnotesize{\textbf{Comparisons on the full multi-stage routing task.} Our full method significantly outperforms both end-to-end (non-hierarchical) baselines and a hierarchical method with a non-learned higher-level state machine.}}
    \label{HighLevelBaselines}
    \vspace{-0.3cm}
\end{table}

We can draw several conclusions from these results: (1) The poor performance of the flat BC and BeT methods is consistent with our hypothesis that hierarchically organized policies are important for recovering from compounding errors over the stages of the task. (2) Methods like ACT, which resemble a kind of implicit hierarchical approach by predicting temporally extended action sequences, still significantly underperform our explicit hierarchical policies. (3) However, designing the high-level state machine manually is also insufficient, and the recovery strategies acquired automatically by our learned higher-level policies lead to more than double the success rate of the hand-designed state machine.

\begin{table}[h]
    \centering
    \begin{tabular}{l|c}\toprule
         Model & Success Rate \\\midrule
         Hierarchical Imitation, no history & 0 / 24 \\
         Hierarchical Imitation (ours) & \textbf{12 / 24} \\\bottomrule
    \end{tabular}
    \caption{\footnotesize{\textbf{History ablation} for the high-level policy, showing that including the history of previously triggered primitives is essential for good performance.}}
    \label{HighLevelAblation}
\end{table}

\noindent \textbf{Ablation experiments with the high-level policy.} Table~\ref{HighLevelAblation} compares the experimental results for our full method with a memoryless variant, demonstrating the importance of using history information. Specifically, withholding the primitive history embedding causes the policy to fail when switching between clips.
After one clip is successfully routed, the memoryless policy continuously recognizes the completion of the routing through the first clip and repeatedly prompts the  \texttt{go\_next} primitive, but fails to move on and begin routing the next clip (i.e., prompting a  \texttt{pickup}). In practice, we found maintaining a reasonable length of context history helps the high-level policy make better decisions; for example, more recovering behavior will be attempted.

\begin{figure}[h]
    \centering
    \setlength\tabcolsep{2pt}
    \begin{tabular}[t]{ccc}
    \begin{subfigure}[b]{0.32\linewidth}
        \includegraphics[width=\linewidth]{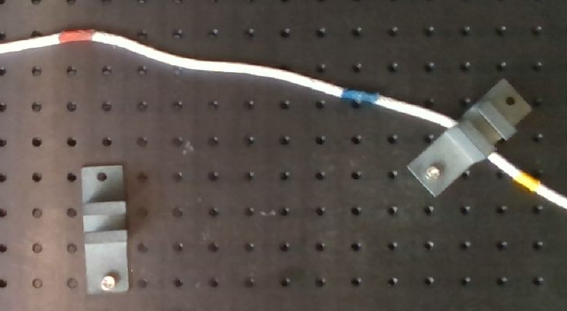}
        \caption{Before Perturb}\label{pertub twice 1}
    \end{subfigure} &  
    \begin{subfigure}[b]{0.32\linewidth}
        \includegraphics[width=\linewidth]{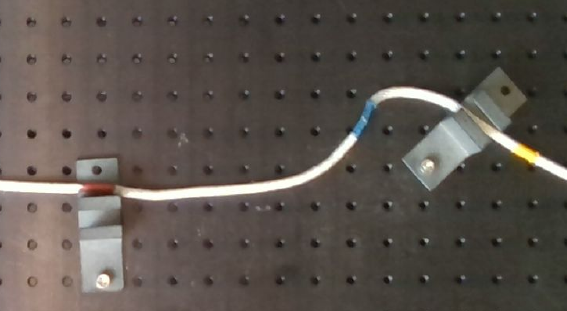}
        \caption{Perturbed Once}\label{pertub twice 2}
    \end{subfigure} & 
    \begin{subfigure}[b]{0.32\linewidth}
        \includegraphics[width=\linewidth]{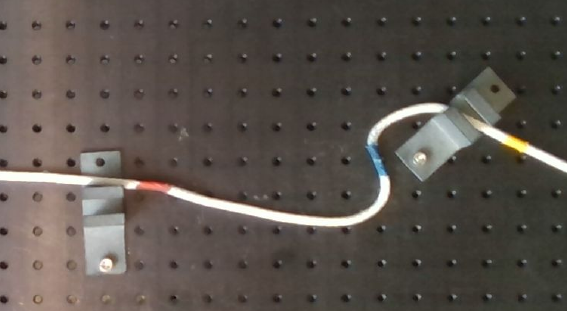}
        \caption{Perturbed Twice}\label{pertub twice 3}
    \end{subfigure} \rule[-1.2ex]{0pt}{0pt}\\
    \end{tabular}
    \caption{This figure shows the \texttt{Perturb\_cable} primitive used twice in a row. \ref{pertub twice 1}: The initial state is hard for the \texttt{Route} primitive to succeed. \ref{pertub twice 2}: The red pickup point on the cable is next to the left clip and cannot be picked up. \ref{pertub twice 3}: The cable is perturbed again so it can be picked up.}
    \label{fig:emerging}
    \vspace{-0.5cm}
\end{figure}

\noindent \textbf{Qualitative analysis of learned behavior.}
Our high-level policy makes decisions by processing its sensory inputs. By training on a diverse dataset end-to-end, our policy in theory should also generate new emergent behavior that was not seen in the training dataset. Indeed, we observed a few interesting behaviors at test time. The robot expanded the usage of the \texttt{Perturb\_cable} primitive combining with other primitives to create novel recovery mechanisms. For example, as shown in Fig.~\ref{fig:emerging}, the robot applies \texttt{Perturb\_cable}. However, the cable resulted in the inner corner of the clip where it couldn't be picked up; then the robot applied again \texttt{Perturb\_cable}  to adjust the shape of the rope to a level it could pick up the cable. 


\noindent \textbf{Failure mode analysis.} We observe that over 95\% of the failures in the system occur when the high-level policy would predict the \texttt{go\_next} primitive while the cable is not actually successfully routed, especially if the cable is just behind the clip, resulting in an irrecoverable state. We hypothesize this is because the visual feature of a clip being inside versus behind the clip is not very distinctive, and perhaps a different camera placement would alleviate the issue.

\subsection{Interactive Fine-Tuning to Quickly Improve Performance}
We additionally demonstrate the ability of our model can fine-tune its performance in both out-of-distribution(OOD) and in-distribution scenarios with a small amount of interactive training, as described in Section~\ref{sec:finetuning}.

\begin{figure}[t]
    \centering
    \includegraphics[width=0.47\textwidth]{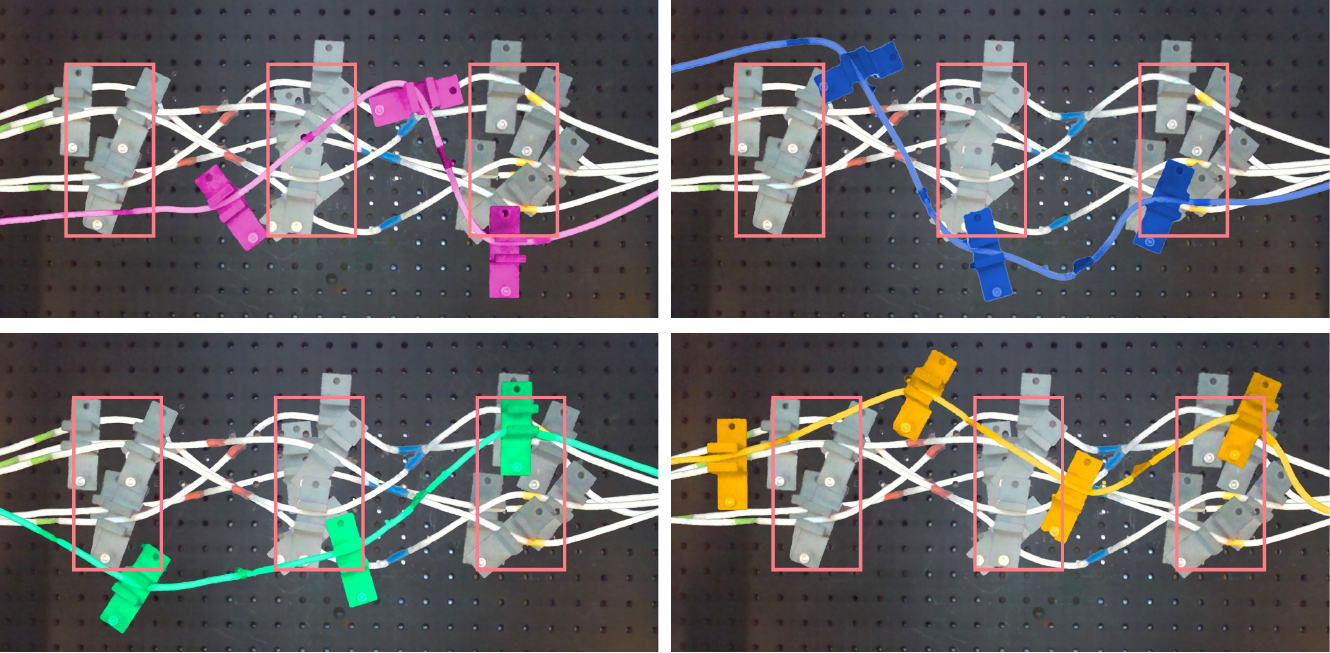}
    \caption{Out-of-distribution clip configurations. \textbf{Top left}, \textbf{Top right}, \textbf{Bottom left}: OOD clip configurations in the 3-clip routing task, the drastically different clip orientation makes them particularly challenging due to the resulting slackness of the cable.   \textbf{Bottom right}: 4-clip routing task with the additional clip not seen in the offline dataset.}
    \label{fig:finetuningimgs}
    \vspace{-0.5cm}
\end{figure}

\begin{figure}[h]
    \centering
        \includegraphics[width=0.5\textwidth]{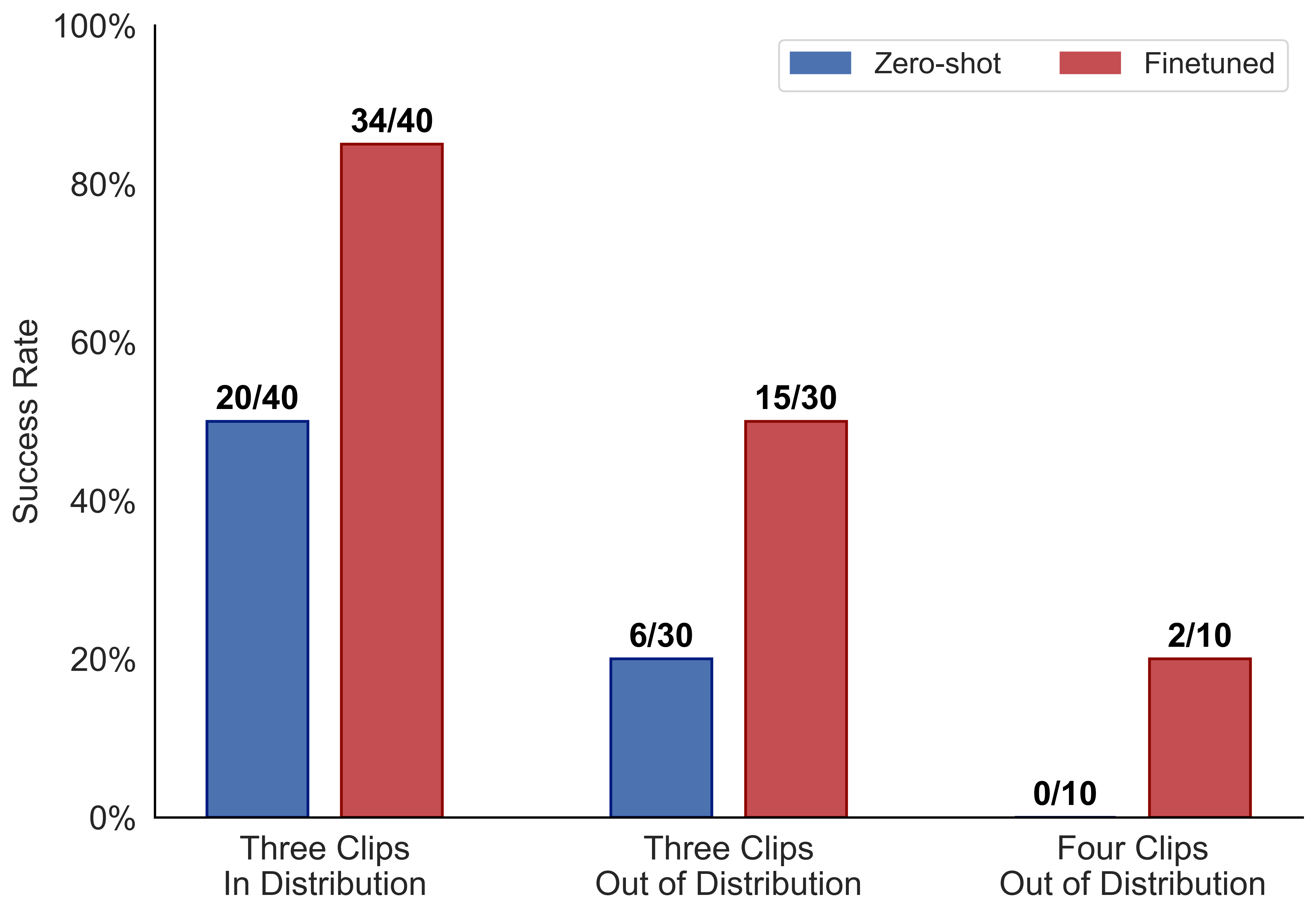}

    \caption{Evaluation of fine-tuning our high-level policy. Fine-tuning enables out-of-distribution generalization to clip configurations in Figure~\ref{fig:finetuningimgs}, including to a configuration where an additional fourth clip was added. This demonstrates the modularity of our system and the ability to extend our method to new clip configurations.}
    \label{fig:finetuningplot}
\end{figure}

\begin{table}[h]
\centering
\begin{tabular}{ l | C{2cm} C{2cm} C{2cm} C{2cm} }
\toprule
\textbf{Configuration} & \textbf{Zero-shot} \newline (10 Trials) & \textbf{Finetuned} \newline (10 Trials) \\
\midrule

\textit{Three Clips In Distribution} \\
\hspace{0.3cm}Configuration 1 & 5 & 9 \\
\hspace{0.3cm}Configuration 2 & 6 & 9 \\
\hspace{0.3cm}Configuration 3 & 4 & 8 \\
\hspace{0.3cm}Configuration 4 & 5 & 8 \\
\vspace{0.1cm}
\hspace{0.3cm}\textbf{Subtotal} & \textbf{20/40} & \textbf{34/40} \\
\midrule

\textit{Three Clips Out of Distribution} \\
\hspace{0.3cm}Configuration 1 & 1 & 3 \\
\hspace{0.3cm}Configuration 2 & 2 & 6 \\
\hspace{0.3cm}Configuration 3 & 3 & 6 \\
\vspace{0.1cm}
\hspace{0.3cm}\textbf{Subtotal} & \textbf{6/30} & \textbf{15/30} \\
\midrule

\textit{Four Clips Out of Distribution} \\
\hspace{0.3cm}Configuration 1 & 0 & 2 \\
\vspace{0.1cm}
\hspace{0.3cm}\textbf{Subtotal} & \textbf{0/10} & \textbf{2/10} \\
\midrule
\hspace{0.3cm}\textbf{Total} & \textbf{26/80} & \textbf{51/80} \\
\bottomrule
\end{tabular}
\caption{Detailed table comparing success rates of zero-shot and fine-tuned high-level policies on 4 in-distribution configurations with three clips, 3 out-of-distribution configurations with three clips, and 1 out-of-distribution configuration with four clips.}
\label{tab:finetuning}
\vspace{-0.5cm}
\end{table}



\noindent \textbf{Out-of-distribution finetuning.} We first evaluate our system on the configurations presented in Figure~\ref{fig:finetuningimgs}. Three configurations (OOD 1, OOD 2, OOD 3) were specifically selected so that some clips were outside the 12.5cm by 15cm box used for selecting the corresponding clip configurations in the training set, or outside the 0 to 45 degree range of orientations used in training, which resulted in scenarios that present a particularly significant generalization challenge. The fourth configuration was selected to include an additional fourth clip that did not appear in any of the original training data for the high-level policy. For each of the four out-of-distribution clip configurations, we collected 10 interactive demonstrations which were used to fine-tune the policy. Figure~\ref{fig:finetuningplot} shows the overall generalization performance of the fined-tuned policies compared with the direct zero-shot transfer of our original policy while Table ~\ref{tab:finetuning} details the number of successes for each configuration. In the case of three clips, we find our policy was able to quickly improve its performance twofold with this handful amount of fine-tuning data. In the four-clip case, due to view shifts observed by the side camera, the original policy was not able to succeed at all. However, the fine-tuned policy was able to rapidly adapt to this new challenging OOD situation with only ten demonstrations.

\noindent \textbf{In distribution finetuning.} It is often of concern in industrial robotic applications that a task should be executed with a fairly high success rate to imply the potential deployment of such systems. Towards that end, we further study how our system can improve reliability for a particular case when the initial success rate is not sufficiently high. To study this setting, we finetuned our pre-trained model on four new randomly sampled clip configurations within its training distributions (separate from the configuration in Section VI.C). Presented in Table ~\ref{tab:finetuning}, our system was able to improve its performance from 50\% to 85\% with only ten additional demonstrations each measured over 10 trials per each of the four configurations for a total of 40 trials. This suggests our system not only enjoys broad generalization capability but also rapidly improves its reliability when a high success rate is desirable.

\label{experiments}
\section{Discussion and Future Work}

We presented a hierarchical imitation learning system applied to the task of cable routing. Our approach is based on the principle that temporally extended multi-stage tasks become more practical when each stage of the pipeline can compensate for and correct mistakes when they arise. In this way, performing a task that requires multiple stages (e.g., inserting the cable into a series of clips) does not lead to a success rate that drops exponentially with each step. The high-level policy in our system, which selects primitives at each stage, can trigger primitives to retry or correct mistakes, and the learned low-level clip insertion primitive can also correct small mistakes because of the corrections present in the demonstrations. In our experimental evaluation, we show that this approach enables the robot to route cables through a series of clips, and even for harder clip arrangements where the system does not succeed consistently, it can be fine-tuned with as few as 10 additional trials. 

Our method still has a number of limitations. Although the success rate of our approach significantly exceeds that of baselines that do not employ learned policies at both levels of the hierarchy, the absolute success rate is still not perfect for industrially relevant applications. Of course, as with all learning systems, larger datasets are likely to lead to improvements in performance. However, it would also be interesting in future work to explore how the addition of more diverse primitives can further enhance the capability of the higher level to correct for mistakes and further reduce failure rates, or how autonomous improvement with RL can improve the method further.
\label{conclusions}
{\appendix
\subsection{View-invariant coordinate system}\label{FirstAppendix}
We use a coordinate system attached to the robot's end-effector frame to express observation and actions. This way, the policy will not over-fit to any particular absolute positions; rather generalize to new clip placements if we can keep the same spatial relativity between clips and end-effector. This is also a convenient mechanism for the system to gain robustness by randomizing the end-effector's initial pose; from the policy's perspective, this is equivalent to \textit{physically} moving the goal (clip) but requires no additional mechanical apparatus beyond the robot itself.

Firstly, we use the wrist camera views to train the policies, which are mounted in the end-effector so they directly enjoy the benefit of the mentioned view-invariant coordinate. We also only use the z-position, or the height, of the end-effector in the policy observation space, which is rotational invariant for our 4DoF action space (translation about the XYZ-axis and rotation about the Z-axis).

The policy outputs the 4D twist $\mathcal{V}_t$ (all translation components, rotation around the z-axis) w.r.t. current end-effector frame $T_t$. For controlling the Franka robot with its control software, we convert it back to the robot's base frame as $\mathcal{V}_t^{\prime}$ using Adjoint mapping, which is a function of $T_t$. Note $T_t$ is a homogeneous transformation matrix as: 
\begin{equation*}
  T_t =   \begin{bmatrix}
R_t & p_t \\
0_{1 \times 3} & 1 
\end{bmatrix}.
\end{equation*}

Then the Adjoint map to relate the two twists in the current frame and base frame can be defined as:
\begin{equation*}
  \text{Ad}_t =   \begin{bmatrix}
R_t & 0 \\
[p_t]\cdot R_t & R_t 
\end{bmatrix},
\end{equation*}
where $[p_t]$ is the skew-symmetric matrix constructed from $p_t$; then we calculate $\mathcal{V}_t^{\prime} = [\text{Ad}_t] \mathcal{V}_t$ which to be sent to the Franka controller. 

\subsection{Neural Network Details}\label{SecondAppendix}
For all neural networks, we use ResNet18 as backbones \cite{he2016deep}. 
For low-level routing policy, we first linear project the z-value to a 128-dimensional vector, then concatenate with ResNet embeddings; then go through a 2-layer MLP with 256 nodes each, and finally output the mean and variance of a TanhGaussian policy. 
For the high-level primitive selection policy, we take the pre-trained ResNet18 from training the low-level policy and keep their parameters frozen while training high-level policy. We pass through three view images through the ResNets to get three embedding vectors, we then up-project the z-value to a 128-dimensional vector. For the primitive sequence embedding, we use a learned word embedding layer \cite{word2vec} to project it to an embedding matrix of size $6 \times 4$, we then flatten this matrix and up-project it to a 128-dimensional vector; we then concatenate all the vectors so far to pass through a single-layer MLP of size 256 with softmax activation in the end.

\subsection{Training Hyperparameters}\label{appendix3}
\begin{table}[H]
    \centering
    \begin{tabular}{l|c}\toprule
         Hyperparameters & Routing Policy \\\midrule
         Optimizer & Adam \\
         Base learning rate & 3e-4 \\
         Weight decay & 0.05 \\
         Optimizer momentum & $\beta_1=0.9$, $\beta_2=0.99$ \\
         Batch size & 512 \\
         Learning rate schedule & cosine decay\\ \bottomrule
         
    \end{tabular}
    \caption{Hyperparameters for training low-level routing policies}
  
\end{table}

\begin{table}[H]
    \centering
    \begin{tabular}{l|c}\toprule
         Hyperparameters & Primitive Selection Policy \\\midrule
         Optimizer & Adam \\
         Base learning rate & 3e-4 \\
         Weight decay & 0.05 \\
         Optimizer momentum & $\beta_1=0.9$, $\beta_2=0.99$ \\
         Batch size & 128 \\
         Learning rate schedule & cosine decay\\ 
         (Finetuning) Warmup Epochs & 5 \\\bottomrule
         
    \end{tabular}
    \caption{Hyperparameters for training high-level primitive selection policies}
  
\end{table}

\begin{table}[H]
    \centering
  \begin{tabular}{l|c}\toprule
         Hyperparameters & Behavior Transformer \\\midrule
         Optimizer & AdamW \\
         Base learning rate & 1e-5 \\
         Weight decay & 2e-4 \\
         Optimizer momentum & $\beta_1=0.9$, $\beta_2=0.99$ \\
         Batch size & 16 \\
         Number of bins k & 64 \\
         Attention Heads & 8 \\
         Block Size & 144 \\
         Decoder Layers & 6 \\
         Output Embedding Size & 256 \\
         Resnet Embedding Size & 512 \\ \bottomrule
         
    \end{tabular}
    \caption{Hyperparameters for training Behavior Transformer}
  
\end{table}

\begin{table}[H]
    \centering
  \begin{tabular}{l|c}\toprule
         Hyperparameters & Action Chunking \\\midrule
         Optimizer & AdamW \\
         Base learning rate & 1e-3 \\
         Weight decay & 3e-3 \\
         Batch size & 256 \\
         Chunk Size & 5 \\
         Optimizer momentum & $\beta_1=0.9$, $\beta_2=0.99$ \\
         Exponential moving average weight & 0.01 \\
         Learning rate schedule & cosine decay\\ \bottomrule

    \end{tabular}
    \caption{Hyperparameters for training with Action Chunking}
  
\end{table}
}\label{append}
\section*{Acknowledgments}
This work was partially supported by ONR N00014-20-1-2383, NSF IIS-2150826, AFOSR FA9550-22-1-0273, and Intrinsic Innovation LLC. We also thank the computing resources provided by the Berkeley Research Computing (BRC) program.\label{ack}

{
\small
\bibliographystyle{plainnat}
\bibliography{refs.bib}
}

\end{document}